\documentclass[10pt,journal,compsoc]{article}

\usepackage[latin1]{inputenc}
\usepackage[american]{babel}     
\usepackage[T1]{fontenc}
\usepackage{amssymb}
\usepackage{subcaption}
\usepackage{graphicx}
\usepackage{longtable}
\usepackage[export]{adjustbox}
\usepackage{xcolor}
\usepackage{textcomp}
\usepackage{algorithm}
\usepackage[noend]{algpseudocode}
\usepackage{amsmath}
\usepackage{amsthm}
\usepackage[singlelinecheck=false]{caption}

\newcommand{\N}{\mathbb{N}}

\newtheorem*{remark}{Remark}

\DeclareMathOperator*{\argmin}{arg\,min}
\DeclareMathOperator\supp{supp}

\usepackage[nocompress]{cite}

\begin{document}

\title{ Inpainting in discrete Sobolev spaces: structural information for uncertainty reduction}

%
%

\author{{\bf Marco Seracini$^a$},\hskip0.4cm {\bf Stephen R. Brown$^ {b,c}$}     \\    \\
 $^a$ Department of Physics and Astronomy, University of Bologna\\
 Via Berti Pichat 6/2, 40127, Bologna, Italy\\ \\
$^b$ Massachusetts Institute of Technology \\
77 Massachusetts Ave, Cambridge, MA 02139, USA\\ \\
$^c$ Aprovechar Lab L3C\\ Montpelier, Vermont, USA \\ \\
{\small {\tt marco.seracini2@unibo.it} \hskip0.5cm 
 {\tt sbrown@mit.edu}} }

\maketitle

\begin{abstract}
In this article, using an exemplar-based approach, we investigate the inpainting problem, introducing a new mathematical functional, whose minimization determines the quality of the reconstructions. The new functional expression takes into account of finite differences terms, in a similar fashion to what happens in the theoretical Sobolev spaces. Moreover, we introduce a new priority index to determine the scanning order of the points to inpaint, prioritizing the uncertainty reduction in the choice. The achieved results highlight important theoretical-connected aspects of the inpainting by patch procedure.

\end{abstract}

 {\bf Keywords:}  Texture Synthesis, Inpainting by Patch, Image Processing, Discrete Sobolev Spaces

\section{Introduction}\label{introduction}

In the literature, the so called ``inpainting problem'' consists in filling zones of ``missing'' information in an image, i.e., in a two-dimensional discrete set \cite{MM}. 

Focusing on the specific two dimensional image case, to have satisfactory results, the inpainted area has to be not ``eye-distinguishable'' from the rest of the given image by a human observer. This subjective way to evaluate the achieved results does not take into account the ground truth, in all of those cases when the to-be-inpainted areas are created deleting zones of the complete starting image \footnote{Even if in practical cases ground truth images are never available, it is possible to create specific tests where the area to reconstruct is expected to be exactly the part that has been deleted from the pristine image. This kind of test is useful, e.g., to verify the correct behavior of the reconstruction algorithms.}.

	To tackle the inpainting problem different solutions have been proposed. On one hand, Partial Differential Equations (PDE) based approaches have been used. These methods involve the introduction of derivatives in the mathematical models (or their equivalent finite differences in the discrete case), working in the so called Sobolev spaces (\cite{BSCB,BBS,CS2,CS,ES,BS}), i.e., in mathematical spaces where the norm of the Lebesgue spaces is extended to the derivatives of the function itself. PDE-based approaches achieve good results when the damaged regions are small in size, otherwise their reconstructions appear blurred. Moreover, these methods usually need a considerable execution time. 

	On the other hand, an interesting promising technique was introduced by \cite{EFROS,WEI}: for each missing point in the to-be-inpainted area, a suitable neighborhood of the point is scrolled all over the known image data and the correct value is chosen maximizing a similarity measure between the neighborhood and the known signal (see, e.g., \cite{PS,EFROS,WEI,ASH,EFROS2,HER,DEM,VESE,GAT}). The quality of the reconstructions by this method is conditioned by the choice of an appropriate similarity measure. Furthermore, despite its conceptual simplicity, this method provides very good results in a relatively affordable execution time (\cite{GG,NN,W}). The inpainting methods based on this idea are referred as "inpainting-by-patch" or exemplar-based techniques.

	In \cite{DEM}, a mathematical formalization describing the behavior of the algorithm introduced in \cite{EFROS,WEI} has been provided through a functional, named ``Inpainting Energy'' ($E_I$). The logic behind $E_I$  is to quantify the difference between a neighborhood of the missing points (reconstructed one by one) and the known part of the signal itself. Another even earlier work (see \cite{BARNEA}), following in some sense a similar approach, used the correlation function in place of $E_I$: here the problem was the rigid registration of two similar images and the goal was the maximization of their similarity.
However, we will show in Sec. \ref{Neighborhood} that  the ``deterministic'' $E_I$ hides a strong probabilistic character: it is equivalent to find, in the given image, which point has the highest probability to match the chosen neighborhood, considering a content-driven, patch-based, similarity.

	Both PDE-based methods and exemplar-based ones have the common goal, hidden by distinct mathematical formulations, to reduce the uncertainty in the choice of the missing points. In the first case, boundary conditions together with constraints on the structure \footnote{From here on, when we will use the word \textit{structure}, we will refer to the  \textit{shape} of the three dimensional structure of the image function (in effect a surface immersed in a 3D space), and not to the geometry of the isophotes, as usually done in the image inpainting contest.} of the unknown are needed (e.g., in \cite{BSCB,BVSO}, the discrete Laplacian is employed as a smoothness estimator); in the second case, the choice is driven by the best match among the available patches in the dataset, usually represented by the given part of the image (in the following sections we will refer to it as to the \textit{training set $T_S$}, borrowing this expression from the Neural Networks literature).

	Despite the goodness of the achieved results, inpainting-by-patch methods mask a conceptual weakness due to neglecting the under-the-surface structural information hidden in $T_S$. This structural information is connected with the concept of finite differences (\cite{MH,MS,ROF}). 
For this reason, introducing a similarity metric that cares about these structural aspects gives support to a better understanding of the logic behind the inpainting process. 
	
	Sobolev spaces (\cite{BRE}), in their discrete versions, have been used in the image processing field with different formalization and aims (see, e.g., \cite{WILSON, DIGESU}). 
 
	We employ, in the context of image inpainting, the definition of Sobolev spaces to formulate a new, neighborhood dependent functional, described in Sec. \ref{Functional Formalization}: in our method we stay true to the patch-based idea, in a way that is quite different from classical PDE methods, in which higher order derivatives are employed for regularization reasons.
	More theoretically, in the seminal paper \cite{ROF}, a similar approach for noise removal has been introduced in BV spaces, giving origin to a vast literature. Differently from \cite{ROF}, in our case, the finite differences are used as a similarity metric, as done in \cite{HSD} for textural classification purposes. An attempt in this direction has been described in \cite{AFCS}, where a quite complete and general variational framework, applyable to super resolution and denoising problems, has been proposed. Our work has points of contact with what developed in \cite{AFCS} and, for the same reasons and in a similar way, our results can be extended to other tasks, as image completion, texture synthesis, image registration (only to cite some), thus providing general operational principles.

\vskip0.15cm
	Another open debate for the solution of the inpainting problem, using the inpainting-by-patch methods, is focused on the scanning order of the missing points. The role of the scanning order is to inpaint first those points where the uncertainty is lower respect to the others. The pivotal work exploiting the importance of the scanning order is represented by \cite{CPT}, in which a priority index has been introduced. The proposed procedure gave considerable results such that it is still commonly taken as reference for qualitative results evaluations. For this reason, in what follows, we have compared our reconstructions to the ones achieved using \cite{CPT}. Starting from \cite{CPT}, further attempts to improve the final quality have been done by different authors (see, e.g., \cite{XWL}), basing their works mostly on numerical considerations. Differently from them, we reformulate a completely new priority index, justifying its introduction by uncertainty reduction motivations.
\vskip0.15cm
	An important scenario, for the inpainting problem, is represented by the fast and effective  Neural Networks (NN) solutions. Recently, many inpainting methods have been developed based on Deep Learning techniques (see, e.g., \cite{RFB,KRJI,WW}). The advantages of NN approaches reside in a fast computational time (after the time consuming initial training) with a high rate of success. Given these premises, the attempt to overpass the NN results appears an insuperable task. The only critic one can move to NN is their well known lack of complete theoretical foundations, such that their design  is essentially nowadays still driven by heuristics. On one hand, the aim of the work presented in this article is to investigate and unveil some hidden aspects of the deterministic inpainting procedure. On the other hand, this may  guide the development of fast machine learning algorithms that have solid and understandable theory and provide benchmarks for comparison. Finally, we auspicate that some aspects considered in this work could be included in the formulation of new NN frameworks.
\vskip0.15cm
	In what follows, the new functional formalization is given in Sec. \ref{Functional Formalization}, probabilistic considerations about the neighborhood are given in Sec.  \ref{Neighborhood}, the concept of causality is exploited in Sec. \ref{Causality}, the importance of the uncertainty and scanning order is investigated in Sec. \ref{Uncertainty}, the descriptions of the numerical implementation adopted and of the role of the training set are in in Sec. \ref{Numerical Implementation} and Sec. \ref{Training set} respectively. The achieved results are shown in Sec. \ref{Numerical Results}. Final remarks conclude the paper in Sec. \ref{Conclusions}. To improve the readability, in the Appendix at the end of the work a list of used symbols and their meanings is provided.

\section{Functional Formalization} \label{Functional Formalization}

\begin{figure} 
\begin{center} 
\includegraphics[scale=0.5]{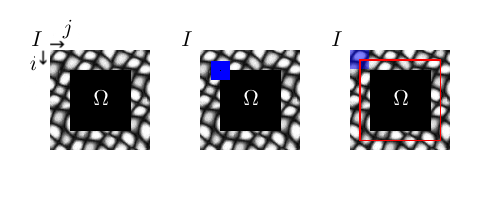}
\caption{\footnotesize On the left the image $I$ contianing the to-be-inpainted area $\Omega$; in the center, the neighborhood $\chi$ in blue, centered in the top-left point to be inpainted; on the right, the red line represents the set $T_S$ of points for which $E_{C_{T}}$ is well defined.}
\label{fig:I}
\end{center} 
\end{figure}

Working with a W-bits coded image function $I$, of size $M \times N$ ( $M$ rows, $N$ columns), containing the area $\Omega$  to be inpainted (see Fig. \ref{fig:I}), we consider a function $\chi : [1,2L+1]^2 \subset \N^2 \rightarrow [0, 2^W-1]$ ($L\in \N$ determines the size of the square $\supp \chi$), that, centered in the generic point $(i,j)$ of $I$, is defined as

\begin{equation*}
\chi_{i,j}:=\chi(i+a,j+b)=I(i+a,j+b)
\label{eq:chi}
\end{equation*}

\hskip-0.5cm with $a=\{-L,-L+1,...,L-1,L\}$, $b=\{-L,-L+1,...,L-1,L\}$, $i \in [L+1,M-L]$ and $j \in [L+1,N-L]$ \footnote{We assume the first element of a matrix to be at coordinates (1,1).}.

We will refer to $\supp \chi$ in the following as the neighborhood of the point to be inpainted (see blue square in Fig. \ref{fig:I} again).	            
The choice as of a square-shaped support for $\chi$, with an odd value for its dimensions, simplifies the mathematical notation without loss of generality \footnote{On the contrary, this aspect must be taken into account in the numerical implementation.}. We will see in Sec.  \ref{Neighborhood} how the shape and size of $\chi$ (i.e., value of $L$) is fundamental for the correct reconstruction of $\Omega$. 

	We outilne the inpainting task in three main sub problems:
\begin{enumerate}
\item identifying the best \textit{content-driven} solutions;
\item identifying the best \textit{structure-driven} solutions;
\item inpainting using the best possible value, resulting from the combination of the best fits  individuated in the two previous steps.
\end{enumerate}

 In general, it is possible to define a pointwise distance function between two points of coordinates  $(i,j)$ and $(h,m)$ in $I$ as
\begin{equation}
d(I(i,j),I(h,m)):=||I(i,j)-I(h,m)||\;,
\label{eq:d}
\end{equation}

\hskip-0.5cm where $||\:.\:||$ is a generic norm.
	Using a similar approach as the one described in \cite{DEM}, the best \textit{content-driven} correspondence for $\Omega$ is related to the minimization of the functional 
$$
E_{C_{T}}(r,s):=\sum_{(i,j)\in \Omega} E_C((i,j),(r,s)),
$$
where, given a point of coordinates $(r,s)\in I\setminus\Omega$, $E_C((i,j),(r,s))$ is defined as
\begin{equation}
E_C((i,j),(r,s)):=\sum \limits_{\substack{(p,q)\in \supp \chi_{i,j}\setminus(i,j)\\ (h,m) \in \supp \chi_{r,s}\setminus(r,s)}} d(I(p,q),I(h,m))\;.
\label{eq:EC}
\end{equation}
The coordinates $(r,s)$ varies in a suitable subset $T_S$ of $[1,M] \times [1,N]$, such that the above sum is well defined without the introduction of further boundary conditions (see red zone in Fig.\ref{fig:I}).
In eq. (\ref{eq:d}), the $||\:.\:||_2$ norm, originally used in \cite{DEM}, has been replaced by a generic norm $||\:.\:||$. We will explicit this norm in the following.

To the light of the previous consideration, the functional $E_{C_{T}}$ represents a sort of \textit{content-related} energy: it is the amount of error one commits comparing two patches having same support, equal to $\supp \chi$.

	In a fashion similar to the one used to formalize $E_{C_{T}}$, we introduce a new \textit{structure-driven} functional, defined as follows
$$
E_{S_{T}}((r,s),l):=\sum_{(i,j)\in \Omega} E_S((i,j),(r,s,),l),
$$
where 

\begin{equation}
\label{eq:E_S}
\begin{split}
&E_S((i,j),(r,s),l):=\\
&\sum\limits_{\substack{(p,q)\in \supp \chi_{i,j}\setminus(i,j)\\ (h,m) \in \supp \chi_{r,s}\setminus(r,s)}} \sum_{k=1}^{l} \frac{1}{R_k} \sum_{\theta \in \Theta_k} d(\Delta_{\theta}^{(k)} I(p,q),\Delta_{\theta}^{(k)} I(h,k)).
\end{split}
\end{equation}

\begin{figure} 
\begin{center} 
\includegraphics[scale=0.3]{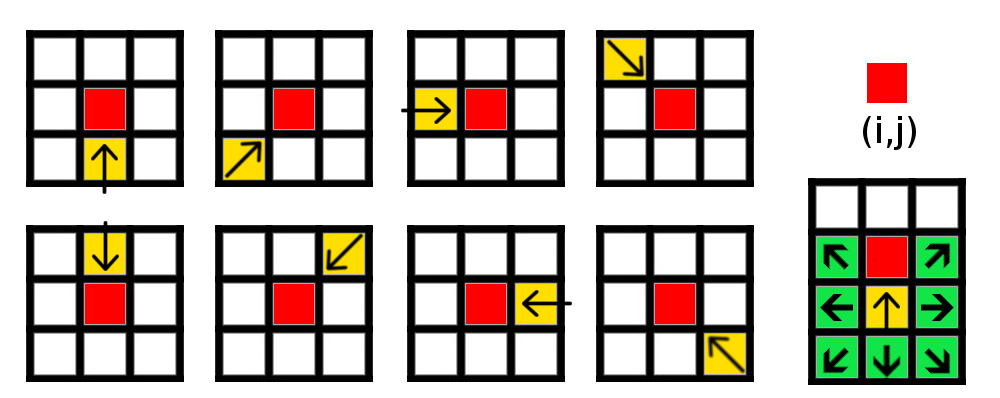}
\caption{\footnotesize Example of not valid directions for $k=1$. The red square is the missing point to-be-inpainted, having coordinates $(i,j)$; the grid represents $\supp \chi$ (being $3 \times 3$ in size for finite differences of order one); the arrows, connecting the yellow squares with the red ones, identify the not valid directions: being centered in the yellow squares it is not possible to calculate the corresponding finite difference of order one  because the value of the red square is unknown. It follows that, for each point, 7 directions are valid, and they are identified by the lines connecting the center of the white squares with the center of the red one. Indeed, in this example, $\Theta_1=\{ 0^{\circ}, 45^{\circ}, 90^{\circ} ,135^{\circ}, 180^{\circ}, 225^{\circ}, 270^{\circ}, 315^{\circ} \}$, $\#\Theta_1=8$, $R_{1}=56$.}
\label{fig:directions}
\end{center} 
\end{figure}

In the above formula, $k$ is the order of the finite differences $\Delta_{\theta}^{(k)} I(p,q)$ and $\theta$ is their orientation, given the set $\Theta_k$ of all the available and \textit{valid} orientations in the considered discrete set. A direction is considered \textit{valid} if the calculation of the associated finite difference $\Delta_{\theta}^{(k)} I(p,q)$ does not involve the to-be-inpainted point of coordinates $(i,j)$. In the standard inpaint-by-patch process the value of $I$ in $(i,j)$ is excluded from the calculation of the pointwise differences: in the same way we excluded it in the finite differences calculation. In Fig. $\ref{fig:directions}$ an example of the not valid directions for the calculation of the finite differences of order one is given: the directions highlighted by the black arrows, connecting the red squares with the yellow ones are not valid, being $I(i,j)$ the unknown. The normalization coefficients are $R_k=8k(\# \Theta_k-1)$ with $\#\Theta_k=8k$, where the symbol $\#$ denotes the cardinality of a set. The maximum order of finite differences $l$ is chosen accordingly with the size of $I$ and of $T_S$. 

	The finite differences in the direction $\theta=0$, from order 1 to $k$,  are defined as
\vskip0.2cm
 $
 \Delta_{0}^{(1)} I(p,q)=I(p,q+1)-I(p,q);
 $
 \vskip0.1cm
 $
 \Delta_{0}^{(2)} I(p,q)=\Delta_{0}^{(1)}  \Delta_{0}^{(1)} I(p,q)=I(p,q+2)+I(p,q)-2I(p,q+1);
 $
 \vskip0.1cm
 ...
 \vskip0.1cm
 $
 \Delta_{0}^{(k)} I(p,q)= \underbrace{ \Delta_{0}^{(1)} ... \Delta_{0}^{(1)}}_\text{k times} I(p,q).
 $
\vskip0.2cm

We recall that, being in a discrete case, the finite differences have a local (and not pointwise like the derivatives) character, as evident from the definitions above.

	The logic paving the definition of $E_{S_{T}}$ is the same of $E_{C_{T}}$, except that the functional calculation involves, here, the finite differences for each available direction in the digital image set. For this reason, similarly to $E_{C_{T}}$, we interpret the functional $E_{S_{T}}$ as representative of a \textit{structure-associated} energy, it carrying information connected with the structure of the image content. 
The reason we employ the word "structure" stands in the fact that derivatives are representative of the shape, i.e., the ``structure'' of a function. Knowing the tangent to a function in each of its points is equivalent to know its structure, the same way the knowledge of an ordinary first order differential equation is effective for the knowledge of the unknown function, given the initial conditions. 

	The \textit{content-related} term considers the pointwise differences only, while the \textit{structure-associated} term takes into account of the local variations contribution.

Combining $E_{C_{T}}$ with $E_{S_{T}}$ in a new functional $E_T:= E_{C_{T}}+E_{S_{T}}$, we can control the contributions related to both the content and the structure in the inpainting process. Being $E_C$ and $E_S$ dimensionally different, the introduction of normalization coefficients $\beta_k$, is needed. The expression of $E_S$ is, then
\begin{equation*}
\label{eq:Esk}
\begin{split}
& \hskip3.5cm E_S((i,j),(r,s),l):=\\
&\sum\limits_{\substack{(p,q)\in \supp \chi_{i,j}\setminus(i,j)\\ (h,m) \in \supp \chi_{r,s}\setminus(r,s)}} \sum_{k=1}^{l} \frac{\beta_k}{R_k} \sum_{\theta \in \Theta_k} d(\Delta_{\theta}^{(k)} I(p,q),\Delta_{\theta}^{(k)} I(h,k)),
\end{split}
\end{equation*}
with $\beta_k \in \mathbb{R}$ (their values will be calculated in Sec. \ref{Neighborhood} ).
If, and only if, $\chi$ is causal \footnote{According to the definition of causality in Signal Theory, a filter is defined causal if and only if the unknown value depends only on the ones given in the past. Strictly speaking, in the 2D asset, this means that only one single point, for each $\chi$, should be unknown (but this is quite never the concrete case). In Sec. \ref{Causality} we will clarify this point.}, then the minimization of  $E_T$ gives the best choice for $I(i,j)$,
$$
I(i,j)=I(\argmin\limits_{(r,s)\in T_S}{\{E_T(r,s)\}}),
$$
where, as before, $T_S$ is a suitable set of coordinates such that, now, both $E_{C_{T}}$ and $E_{S_{T}}$ are well defined  \footnote{For simplicity of notation we use, here, the same symbol $T_S$ as in eq. \ref{eq:EC}, even if, in general, the two sets could be not necessarily equal.}. As we will show in Sec. \ref{Neighborhood}, the introduction of $E_{S_{T}}$ reduces the uncertainty connected with the choice of the best possible value in $T_S$, for a missing point in $\Omega$.

\section{Neighborhood} \label{Neighborhood} 

For simplicity of notation, in what follows we do not explicitly indicate the dependency of the energy from $(i,j)$ and $(r,s)$.
The new functional $E_T=E_{C_{T}}+E_{S_{T}}$ hides a probabilistic interpretation. To prove this statement,  we initially focus on $E_{C_{T}}$.
For each point $(i,j) \in \Omega$, $E_C$ represents the sum of the differences, in modulus, in a domain of size $\supp \chi_{i,j}$, between $\Omega$ and the subset $T_S \in I \setminus \Omega$: the closer this sum is to zero, the higher the probability is to find a correct match.

	The value of $E_{C}$ is the sum of the $d((h,m),(p,q))$ contributions, as expressed in eq. (\ref{eq:EC}). We can normalize each $d((h,m),(p,q))$ in $\supp \chi_{i,j}$ as follows
\begin{equation*}  \label{p}
d_{C Norm}((h,m),(p,q)):=1-\frac{d((h,m),(p,q))}{(2^W-1)} \leq 1 
\end{equation*}
such that we can define

\begin{align*} 
\begin{gathered}
E_{P_{C}}((h,m),(p,q)):=\\:=
\begin{cases} 
d_{C Norm}((h,m),(p,q))\:,\: if\: d_{C Norm}\neq 0\;,\\
\varepsilon,\:\:otherwise\\
\end{cases} 
\end{gathered}
\end{align*}

\begin{figure}
\begin{center}
\includegraphics[scale=0.4]{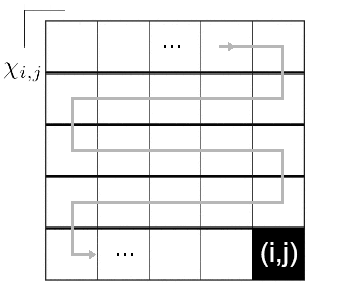}
\end{center}
\caption{\footnotesize Raster scanning order for a causal neighborhood. The black square is the point $(i,j)$ to be inpainted, the grid is the neighborhood $\supp \chi_{i,j}$. The gray path shows the raster scanning order. The definition of causality for a neighborhood assumes that all the values in the white square are known.}
\label{fig:N} 
\end{figure}

\hskip-0.5cm where $\varepsilon$ is arbitrarily small and $E_{P_{C}}((h,m),(p,q))$ is the probability to have a match in $\supp \chi_{i,j}$.
Introducing a scanning order of $\supp \chi_{i,j}$, e.g (without loss of generality), the raster scanning (see Fig. \ref{fig:N}), and switching to a logarithmic scale, the patch matching probability $E_{M_{C}}$ can be expressed as
\begin{gather}
E_{M_{C}}(i,j)= -\prod\limits_{\substack{(p,q)\in \supp \chi_{i,j}\setminus(i,j)\\ (h,m) \in \supp \chi_{r,s}\setminus(r,s)}}log[1-E_{P_{C}}((h,m),(p,q))]\: .
\label{PM}
\end{gather}

If  $E_{M_{C}}(i,j)$ is close to zero (ideally zero for each term of the product) the reconstruction is exact, i.e., without errors of any kind. On the other hand, if $E_{M_{C}}(i,j)$ is high, the probability to have a correct match is very low.

From equation (\ref{PM}) two conditions follow:
\begin{itemize}
\item a hypothesis of statistical independence is hidden in the formulation; 
\item dropping a \textit{not equal to one} term in the product (\ref{PM}) is equivalent to drop part of the information related to the structure of $I$.
\end{itemize}

From the previous considerations it follows that the shape and the size of $\supp \chi_{i,j}$ both determine the quality of the reconstructions. In particular, the second point states that it is not enough to take into account only the two orthogonal principal directions for the reconstruction (as it is usually desiderable in practice), because neglecting a point in $\supp \chi$ is equivalent to assume to know its value with probability equal to one, i.e., that its value does not condition the outcome (size and shape of the neighborhood change the value of the \textit{confidence} term introduced in \cite{CPT}).

When the two above conditions, stating an ``ideal'' case, are not respected, the inpainting process does not generally perform correctly, the quality of the results depending on how far from the ideal case the real case is. 
We stress that $\supp \chi$ is discriminating for content as it is for structure, but its shape and size are not sufficient for a correct reconstruction. In fact, if $\supp \chi$ is made bigger and $E_{M_{C}}(i,j)$ stays close to one, we have a high probability to have made the correct choice. The other way around, if $E_{M_{C}}(i,j)$ decreases when $\supp \chi$ grows, we have individuated a local similarity that is going to be lost increasing the "field of view" (i.e., the $\supp \chi$). From the previous consideration comes that the correct choice of $\supp \chi$ depends on the image content.

\begin{figure}
\begin{center}
\includegraphics[scale=0.82]{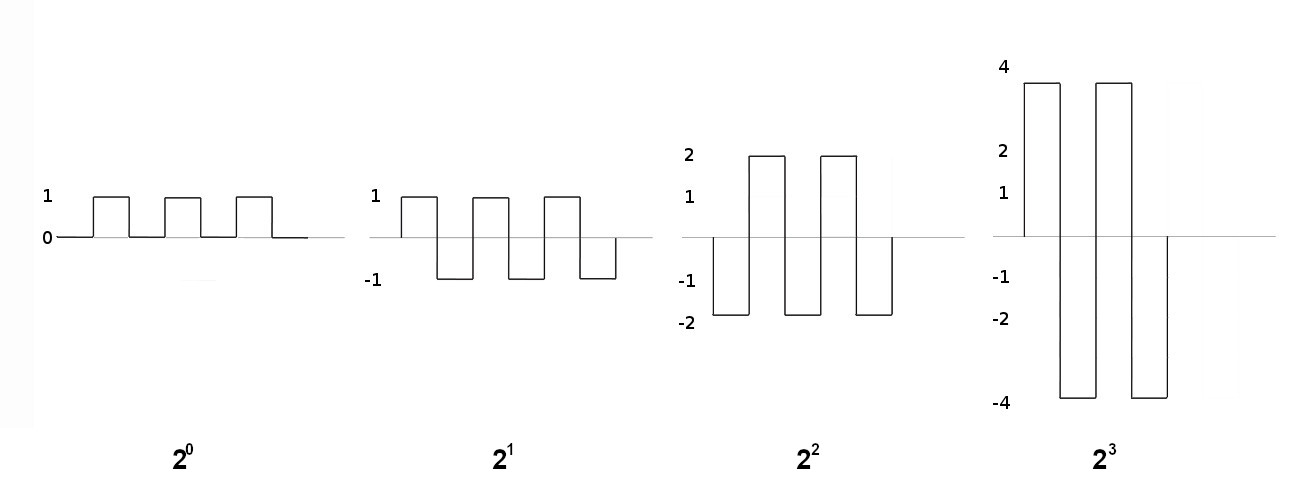}
\end{center}
\caption{\footnotesize A square wave signal, with values in the range $[0,1]$, represents the normalized signal with the highest amount of variations, whose maximum value is equal to $1\:(=2^0)$ (first figure on the left). We assume the values 0 and 1 to be respectively the minimum and the full scale of the measurement instrument. Performing the calculation of the finite difference of first order (second figure from the left) the range of variation expands to $2=2^1$ and so on, proceeding exponentially with the differentiation (last two figures on the right, relative to the second and third order finite differences, respectively).}
\label{fig:beta} 
\end{figure}

	For what concerns $E_S$, in a similar manner used for $E_C$, it is possible to write
\begin{equation*}  
\begin{split}
&d_{S Norm}(\Delta_{\theta}^{(k)} I(p,q),\Delta_{\theta}^{(k)} I(h,m)):=\\
&1-\frac{d(\Delta_{\theta}^{(k)} I(p,q),\Delta_{\theta}^{(k)} I(h,m))}{2^k(2^W-1)} \leq 1 
\end{split}
\end{equation*}
and
\begin{align*} 
\begin{gathered}
E_{P_{S}}((h,m),(p,q)):=\\:=
\begin{cases} 
d_{S Norm}(\Delta_{\theta}^{(k)} I(p,q),\Delta_{\theta}^{(k)} I(h,m))\:,\: if\: d_{S Norm}\neq 0,\\
\varepsilon, otherwise\\
\end{cases} 
\end{gathered}
\end{align*}
where, again, $\varepsilon$ is arbitrarily small and $E_{P_{S}}((h,m),(p,q))$ is the probability to have a match in $\supp \chi_{i,j}$.
Reasoning the same way we did for $E_C$, the patch matching probability $E_{M_{S}}$ can be expressed as
\begin{gather*}
E_{M_{S}}(i,j)= -\prod\limits_{\substack{(p,q)\in \supp \chi_{i,j}\setminus(i,j)\\ (h,m) \in \supp \chi_{r,s}\setminus(r,s)}}log[1-E_{P_{S}}((h,m),(p,q))]\:.
\end{gather*}
\hskip0.5cm Due to the normalization coefficient $2^k(2^W-1)$, the contribution of the finite differences tends to ``refine'' the value of the inpainting energy by a factor $\beta_k\propto\frac{1}{2^k}$. If we assume to work with a bounded function, hypothesis always verified in the applications, the values of $\beta_k$ scale exponentially with base $2$, as evident from Fig. \ref{fig:beta}. The example in figure describes an "extreme" case for which the contributions of subsequent finite differences stay fixed to one. When this does not happen, they tend to decrease as the order of derivation grows.

The structural information is hidden as in $E_{C_{T}}$ as in $E_{S_{T}}$. This consideration enlightens a subtle connection between content ($E_{C_{T}}$) and structure ($E_{S_{T}}$), being the structural information contained as in the size and shape of $\supp \chi$ (i.e., in both $E_{C_{T}}$ and $E_{S_{T}}$), as well as in the finite differences expressions. 

\begin{figure}
\begin{center}
\includegraphics[scale=0.9]{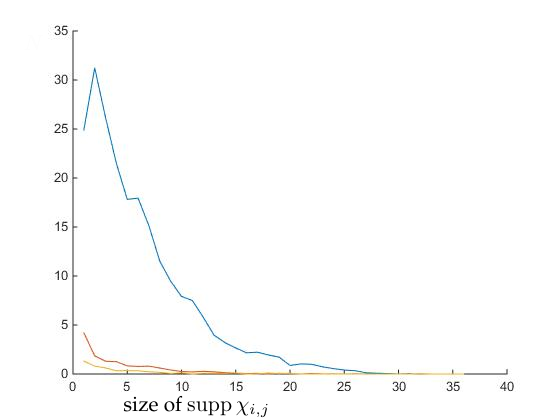}
\hskip0.1cm
\includegraphics[scale=0.9]{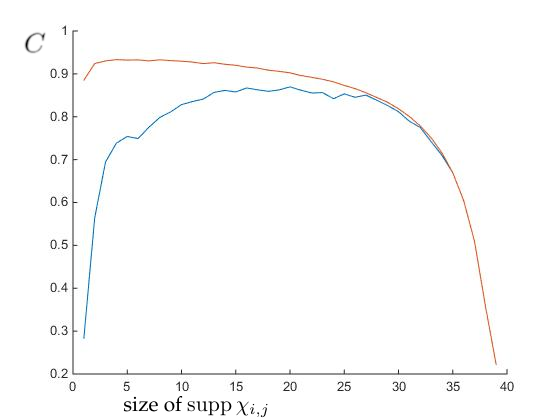}
\end{center}
\caption{\footnotesize On the left, the trend of the number of patches having the same $E_T$ when varying the size (side in pixels) of $\supp \chi$; on the right the trustability $C$ ($=1-U$, where $U$ is the uncertainty) in the choice of the available patches when varying $\supp \chi$. The blue line is relative to the use of $E_{C_{T}}$ only, the red line gives reason to the introduction of $E_{S_{T}}$ (first order differences term in red, both first and second order difference terms in yellow). In particular, closer is $C$ to one, better will be the choice of the value for the reconstruction; on the contrary, lower the number of candidate points, better will be the discrimination between different patches. In the right figure, $C$  (blue line) increases with the size of $\supp \chi$ while tends to decrease when the number of  available patches in the image becomes too small (right side of the graph on the right). Test performed on texture image of size $81 \times 81$ pixels.}
\label{fig:unc} 
\end{figure}

	An application example is given in Fig. \ref{fig:unc}  \label{fig:uncertainity}. Once the shape of $\supp \chi$ has been chosen, the minimization of $E_T$ is equivalent to find the best match between patches: if $\supp \chi$ is too small in size it can occur to find considerably different values associated to the same $E_T$. In substance, the metric used for the calculation of $E_T$ gives origin to a set $S_C$ of equivalence classes, depending on $\supp \chi$. The number of elements in each class decreases as $\supp \chi$ becomes bigger in size as well as including $E_{S_{T}}$. In Fig. \ref{fig:unc}, the effect of the  introduction of $E_{S_{T}}$ on the trend of the trustability $C$ is shown: thanks to it, the reduction of uncertainty $U$, given the same $\supp \chi$ size, is faster. The inclusion of the term $E_{S_{T}}$ decreases $U$ as well as the space of the possible solutions. The role of $E_{S_{T}}$ can be seen as a refinement to discriminate between patches that would otherwise be equally eligible for the inpainting.

\section{Causality} \label{Causality}
\begin{figure}
\begin{center}
\includegraphics[scale=0.1]{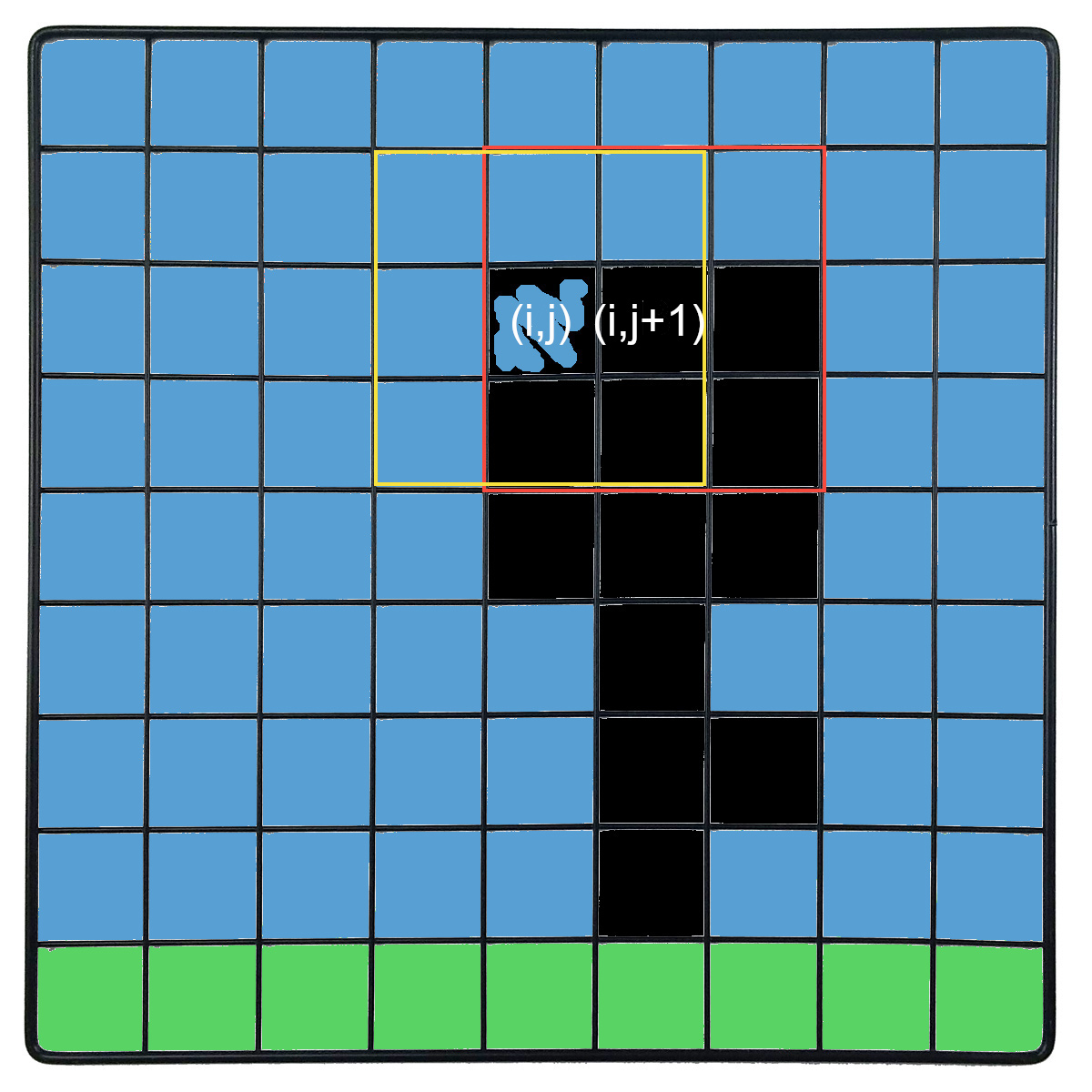}
\end{center}
\caption{\footnotesize The inclusion of unknown points in $\supp \chi$, during the minimization of $E_T$, makes its calculation not effective. In figure, it is assumed to use a square $\supp \chi$, highlighted in yellow when centered in $(i,j)$, i e., at time $t$, and highlighted in red when centered in $(i,j+1)$, i.e., at time $t+1$. In fact, it results $\min\{E_T((i,j),t)\}\neq \min\{E_T((i,j),t+1)\}$}, where the $\min$ operation is performed all over the points  belonging to the not black area.
\label{fig:causality} 
\end{figure}

In Signal Theory, a system is defined to be \textit{causal} if and only if the values of its output depend only on the present and past values of the input (\cite{OWH}). In the inpainting case, we state that the minimization of $E_T$ is meaningful if and only if $\supp \chi$ guarantees the causality condition, i.e., if the calculation of $E_T$ is done including only known points. If this does not happen, the values calculated, lets say in $(i,j+1)$ in Fig. \ref{fig:causality}, modifies the previously calculated ones ($(i,j)$ in the same figure), whose number and position both depend on the size and shape of $\supp \chi$. 

	The inpainting energy $E_T$ changes with time (the points of $\Omega$ are inpainted sequentially), such that the presence of unknown values for its calculation introduces loops in the system (feedback), making the minimization effectively not computable. Moreover, we remark that, in general 
\begin{align*} 
\begin{gathered}
\min \limits_{(r,s)\in T_S} \sum_{(i,j)\in \Omega} E_C((i,j),(r,s))\neq  \\
\neq  \sum_{(i,j)\in \Omega} \min \limits_{(r,s)\in T_S} E_C((i,j),(r,s))\:.
\end{gathered}
\end{align*} 
\hskip0.5cm The previous inequality can be conceptually extended to $E_T$, being $E_{C_{T}}$ just a part of it.

	The procedure described, e.g., in \cite{WEI}, does not take into account this aspect and, to achieve convergence, sets random initial conditions in $\Omega$. Even if a covergence is reached, this way of proceeding has two main drawbacks:
\begin{itemize}
\item the inpainted result is not always the same (it depends on the random initial conditions);
\item if in $T_S$ the missing area is given, such that we expect to find a perfect match for $\Omega$ (i.e., $E_T=0\:\: \forall\:\: (i,j) \in \Omega$), the inpainting procedure does not generally reconstruct it as expected.
\end{itemize}
\hskip0.5cm The missing of the causality condition is a source of uncertainty and its respect is needed to reduce the inpainting error. Two actions can be taken for this scope:
\begin{itemize}
\item considering a variable (in size and shape) $\supp \chi$ to guarantee the causality condition is not violated;
\item defining an appropriate scanning order (priority) that takes into account at the same time of $E_T$, of the uncertainty connected with the shape of $\supp \chi$, and of the isophotes.
\end{itemize}
 All these actions have been taken into account in our work. 

	The first one is mainly algorithmic and consists in considering, in the calculation of $E_T$, only known points in $\Omega$ and in $T_S$ \footnote{A suitable binary mask is enough to numerically implement this condition.}. For this reason from now on, to simplify the notation, and due to the variability of $\supp \chi$, we will refer to the size of $\supp \chi$ to identify the minimum square size in which $\supp \chi$ is contained. 

	For the second action the formalization of a priority index is needed: being it a more complex procedure, we will discuss it in the following Sec. \ref{Uncertainty}.

\section{Uncertainty and Priority} \label{Uncertainty}
The inpainting process consists of two different phases: the \textit{analysis} of the similarity between patches (to determine the best fit) and the \textit{synthesis} of the unknown value. Both of these processes hide sources of uncertainty.
\vskip0.1cm
	In the analysis phase, the scanning order to inpaint $\Omega$ seems to play a central role for the correctness of the reconstruction. In this direction, different works have tried to improve the confidence term firstly introduced in \cite{CPT}, to avoid the too fast convergence to zero of the original formulation (see \cite{CHLWW,XWL,OLPA} ). We shortly recall that in \cite{CPT}, the scanning order (also called priority) $P(i,j)$ has been defined as
\begin{equation*}
P(i,j)=C(i,j)\cdot \frac{D(i,j)}{\alpha}
\label{criminisi}
\end{equation*}
where $(i,j)\in\Omega$, $C(i,j)$ and $D(i,j)$ are, respectively, the confidence term (what we called trustability before), and the data term: the first one estimates the uncertainty of the neighborhood, the second one considers the direction of intersection between the isophotes and the boundary $\partial\Omega$ of $\Omega$.  Briefly,
\begin{equation}
C(i,j)=\frac{\sum\limits_{(r,s) \in \supp \chi \cap \supp I}C(r,s)}{\#( \supp \chi_{i,j})}, \:\:D(i,j)=\frac{| \nabla^{\perp}_{(i,j)}\cdot n_{(i,j)} |}{\alpha}
\label{eq:C}
\end{equation}
where $\#(\supp \chi_{i,j})$ is the cardinality of the pixels in $\supp \chi_{i,j}$,  $\nabla^{\perp}_{(i,j)}$ is the isophote vector and $n_{(i,j)}$ is the unitary vector orthogonal to $\partial\Omega$. $C(i,j)$ is initially set to one in every point of $I\setminus \Omega$ (see \cite{CPT}  for a complete description). 
 The value of $P(i,j)$, being in the discrete, and consequently approximated asset, is strictly connected with the methods used to achieve $ \nabla^{\perp}_{(i,j)}$ and  $n_{(i,j)}$, whose calculation usually requires the introduction of a preprocessing stage by a Gaussian smoothing to reduce noise. Dramatically different reconstructions come out from different implementations of the same method. 

	The contributions of $C(i,j)$ and $D(i,j)$ have been usually considered separated, as respectively explanatory of the uncertainty of the given data and the preferential direction of propagation of the isophotes (\cite{CHLWW,OLPA}). Both $C(i,j)$ and $D(i,j)$ belong to the interval $[0,1]$, given a suitable normalization coefficient $\alpha$ (e.g., 255 if W=8). Exploiting this property, we can interpret them as probabilities. More precisely, $C(i,j)$ is the probability to find a match, given a certain \textit{trustable} number of neighboring points, while, when $D(i,j)$ is normalized, it expresses a probability connected with the structure, i.e., the morphology of $I$ around $\Omega$. 

The last statement follows from the properties of edges in images. In fact, points belonging to contours of an object in an image are those points that generally are less numerous than other points, i.e., that are less probable to be found. 

\begin{figure}
\begin{center}
\includegraphics[scale=0.35]{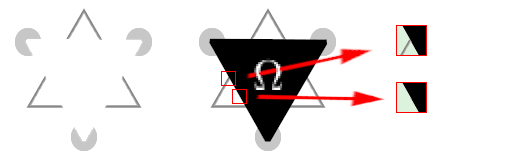}
\end{center}
\caption{\footnotesize On the left the Kanizsa triangle to inpaint; in the center the same tringle with $\Omega$ highlighted in black; on the right two examples of neighborhood, to be inpainted in the black areas. It is easy to verify that the number of possible patches compatible with the top neighborhood is lower than the number of patches compatible with the down one ($2L+1=3$, $l=1$).}
\label{fig:inpaintingKZS} 
\end{figure}

	An explanatory example is provided in Fig. \ref{fig:inpaintingKZS}: to inpaint the Kanizsa (\cite{K})  triangle we have to set the priority.  As an example, we take into account of two plausible configurations of $\chi$, being $\supp \chi$ highlighted in light green in Fig. \ref{fig:inpaintingKZS} and shown in the small squares on the right (in black the missing areas). It is evident that the number of  points, in $I\setminus \Omega$, compatible with the upper square is lower than the number of points compatible with the lower one.  

	Thought in this way, $D(i,j)$ individuates the directions along which there is a reduction of uncertainty in the choice of the best fit. In fact, the equivalence class to which the candidate points belong, has (auspicably) the lowest cardinality respect to the other ones in $S_C$. The probability to make a mistake, in the choice is, then, reduced even in presence of multiple candidates. For this reason, we introduce $D_M(i,j)$, in place of $D(i,j)$, defining it as
\begin{equation}	\label{eq:DM}
D_M(i,j):=\max_{\theta}\{\Delta_{\theta}^{(1)}I(i,j)\}\: .
\end{equation}

Edges are zones where exists a $\Delta_{\theta}^{(1)}$ that is higher than other areas of the image. Then, its maximum individuates the direction of lowest uncertainty. 

	Two main advantages follow from this considerations: 
\begin{itemize}
\item no need to calculate (and approximate) $\nabla^{\perp}$;
\item there are not preferential straight directions of propagation, a known drawback of \cite{CPT}.
\end{itemize}
The reasons we limit the value of $D_M$ to the first order of the finite differences in the definition (\ref{eq:DM}) have numerical and perceptual nature and will be explained in Sec. \ref{Numerical Implementation}.

\vskip0.3cm
Moreover, back to the probabilistic interpretation given to $E_T$ in Sec. \ref{Neighborhood}, its value can be considered a measure of uncertainty itself: where $E_T$ is lower, the uncertainty is lower the same way. Then, we include the value of $E_T$ in the formalization of our new priority index:
$$
P^*(i,j)=[C(i,j)+a^* D_M(i,j)]\cdot b^* E_T(i,j)
$$
where $a^*$ and $b^*$ are two normalization coefficients.

The value of $a^*$ follows from two considerations: $D_M$ is a finite difference, subject to the same normalization reasoning of Sec. \ref{Neighborhood}, and it has to be normalized by the ratio of valid points involved in its calculation, i.e., $C(i,j)$. This last consideration has to be extended to $E_T$ too, providing the value of $b^*$. From these premises follows that, in the case of finite differences of order one (that we used in practical applications), it results $a^*=\frac{C(i,j)}{255}$ and $b^*=C(i,j)$.

	The new $P^*$ balances between the "morphological" uncertainty connected with the shape and size of $\Omega$ and its surroundings (i.e., $[C(i,j)+a^* D_M(i,j)]$), and the uncertainty due to a limited $T_S$ (i.e., $b^*E_T$).

\begin{figure}
\begin{center}
\includegraphics[scale=0.7]{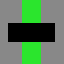}
\hskip0.5cm
\includegraphics[scale=0.35]{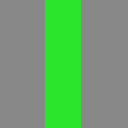}
\end{center}
\caption{\footnotesize On the left the image to inpaint ($\Omega$ in black); on the right the inpainted result ($2L+1=3$, $l=1$). The algorithm guarantees the connectivity principle, whatever is the size of $\Omega$ (result according with \cite{AFCS}).}
\label{fig:inpaintinguncert} 
\end{figure}

\begin{figure}
\begin{center}
\includegraphics[scale=0.5]{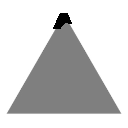}
\hskip1cm
\includegraphics[scale=0.5]{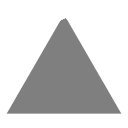}
\end{center}
\caption{\footnotesize On the left the triangle to be inpainted (the black area highlights the missing points); on the right the result of inpainting using $2L+1=3$ and $l=1$. No ``over-shot'' artifacts appear. Moreover, the result is quite symmetric, even if $\Omega$ (in black) is not. The expected top-pointing corner is not achieved because it misses in the known part of the image. The convergence to the stable solution is achieved at the $4^{th}$ iteration.}
\label{fig:triangle} 
\end{figure}

	To quantitatively support our choice, we inpaint a triangular shape (see Fig. \ref{fig:triangle}), the same way they did in figure 8 of \cite{CPT}: in our case, it is possible to see that no ``over-shot'' artifacts appear. The completion of the triangle is not as expected, but this is due to the lack of the right patch in the rest of the known image: no top-pointing corners are available without rotation. 
	In Fig. \ref{fig:inpaintinguncert} the reconstruction by our algorithm of a single straight line shows the connectivity principle is respected.

\vskip0.3cm
	Back to $P(i,j)$, another crucial point for its calculation is represented by the esteem of $C(i,j)$: in eq. (\ref{eq:C}) the known points in $\supp \chi \cap \supp I$ are set to one, independently from $\Omega$. In probabilistic terms this hypothesis is equivalent to state that all the given points in $\supp \chi \cap \supp I$ have the highest \textit{trustability}, i.e., their values are certainly correct. This assumption is true if there is a perfect match between the points to inpaint in $\Omega$ and the available ones in $T_S$, but in the most of the real cases this does not happen. In fact, generally speaking, $E_T(i,j)\neq 0\: \forall (i,j) \in \Omega$ (otherwise the solution of the inpainting problem would be obvious). Moreover, we remind that a reference image unavailable for the quantification of the quality of the reconstructions. 

	From the previous considerations it follows that $E_T(i,j)\neq 0$ has an effect on $C(i,j)$, when $(i,j) \in I\setminus \Omega$, and this effect has to be propagated to the whole image, i.e., the error has to be redistributed. This reasoning is licit only because a reference is missing: only in this case, in fact, if we look at the image from the reference system of the missing points in $\Omega$, we can assume the surrounding points are wrong, viceversa if we look to the problem taking as reference the points in $I\setminus \Omega$. This considerations determine the need to upload $C(i,j)$ also in $I \setminus \Omega$, taking into account that the value of $E_T(i,j)$ is not zero.

\begin{figure}
\begin{center}
\includegraphics[scale=0.2]{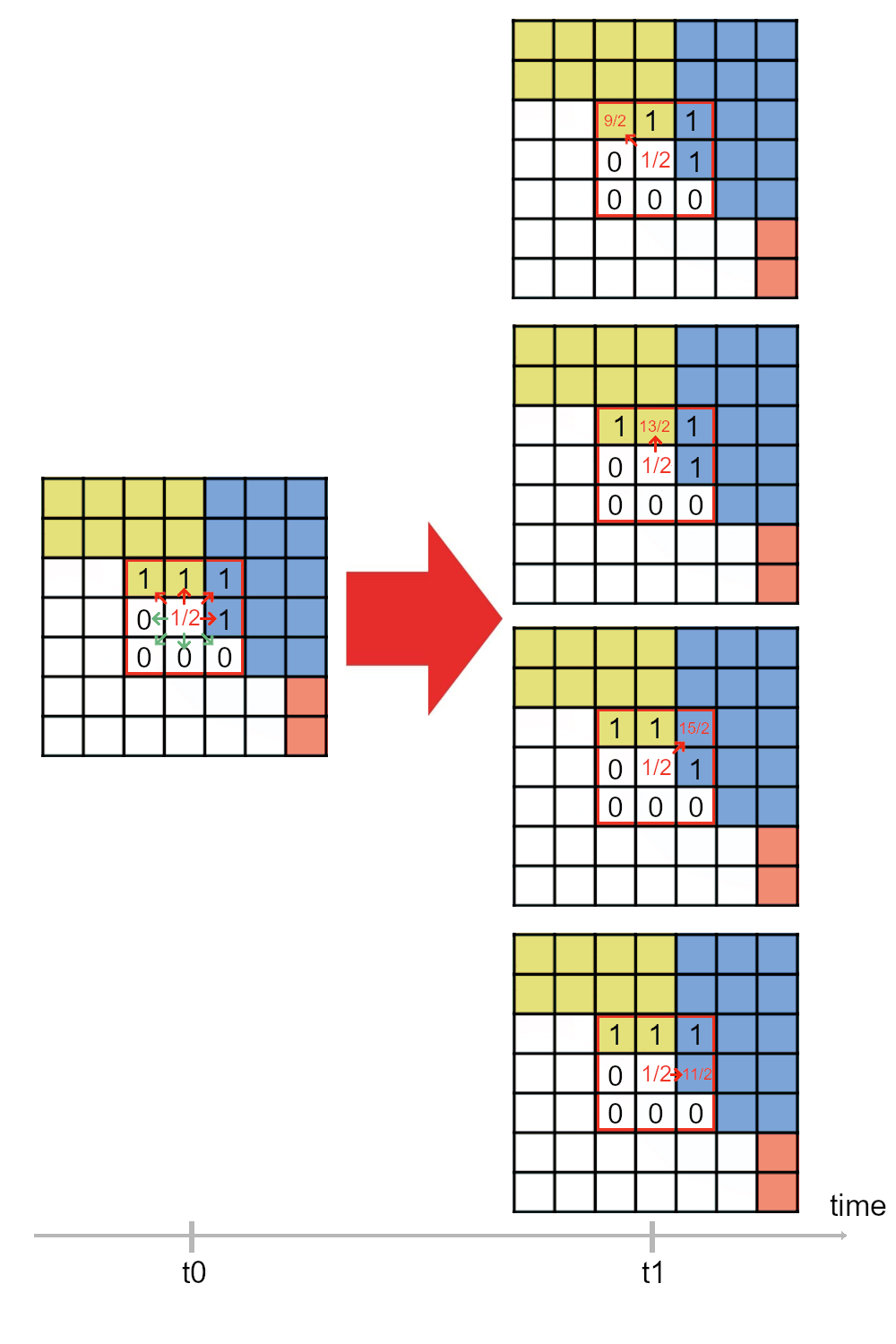}
\end{center}
\caption{\footnotesize Description of the updating process, time arrow from left to right, $\Omega$ in white. On the left, at time $t_0$, the small green and red arrows around the point to inpaint (in the center of the red highlighted square, coordinates $(i,j)$, $C(i,j)=1/2$) determine the directions for the updating of the trustability. On the right, at time $t_1$, the propagation of the new value: while it is tacitly assumed fair to update the values of $C(i,j)$ in $\Omega$ (green arrows, starting zero value), the same has to be done also in $I \setminus \Omega$ (red arrows, starting value equal to one). The linearity allows to update $C(i,j)$ individually for each single point and to sum the results. The Markovian hypothesis limits the radius of influence for the update to the bordering pixels only.}
\label{fig:update} 
\end{figure}

	In Fig. \ref{fig:update} a graphical example is provided.
 We can figure $C$ to be as a field (e.g., of force) where all the points are interconnected: a perturbation occurring somewhere would cause the modification of the entire configuration. To make this update numerically plausible we introduce two hypothesis:

\begin{itemize}
\item the system satisfies the superposition effects principle (linearity);
\item the system is Markovian \cite{GRS}.
\end{itemize}

The first hypothesis allows to update $C(i,j)$ in $I\setminus \Omega$ summing the effects separately; the second one limits the radius of influences of the perturbation to the bordering neighborhoods. 

\begin{figure}
\begin{center}
\includegraphics[scale=1]{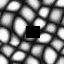}
\hskip0.4cm
\includegraphics[scale=1]{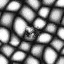}
\hskip0.4cm
\includegraphics[scale=1]{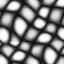}
\end{center}
\caption{\footnotesize On the left the texture to inpaint ($\Omega$ in black). In the center the inpainting result achieved if $C(i,j)$ is not updated in $ I\setminus \Omega$ (i.e., $C(i,j)=1 \forall (i,j)\: \in I\setminus \Omega$, $T_S=I\setminus \Omega$). On the right the inpainting result updating the points in $I\setminus \Omega$ ($2L+1=3$, $l=1$, same $T_S$ as before). The improvement of the result, when correctly updating $C(i,j)$, is evident.}
\label{fig:order} 
\end{figure}

	Neglecting this updating process brings to visual poor results, as shown in Fig. \ref{fig:order}.
\vskip0.1cm
	For what concerns  the synthesis phase, a source of uncertainty is due to the norm used to quantify the patch similarity. In past works it has been stressed how the most appropriate norm for images is the TV norm, that is essentially an $\ell^1$ norm of derivatives \cite{ROF}. On the other hand, in \cite{AFCS}, results achieved using the $\ell^2$ norm are evaluated as ``smoother'' and, for this reason, better than the $\ell^1$-based ones. For this reason, and according with the most of the literature, we have chosen to operate with the $\ell^2$ norm.

\section{Numerical Implementation} \label{Numerical Implementation}

To implement eq. (\ref{eq:E_S}) in a suitable algorithmic form, some considerations are useful to be made.
In the discrete one dimensional case, given a row $r$ of the image $I$, we can rewrite $I(r,i+k), r, k,i \in \mathbb{N}$ as a function of past samples, using the following equation \footnote{In place of a horizontal direction $r$ (i.e., a row), it is possible to consider whatever orientation, at the price to introduce some interpolation methods to compute the missing values on the discrete grid.}:
$I(r,i+k)=I(r,i)+\Delta ^{(1)} I(r,i)+\Delta ^{(1)} I(r,i+1)+ ...+\Delta ^{(1)} I(r,i+k-1)$.
\begin{proof}
From the definition of $\Delta ^{(1)} I(r,i)$, given a generic index $i$, the equality $I(r,i)+\Delta ^{(1)}I(r,i)=I(r,i+1)$ stands.
Then
\vskip0.2cm
$
I(r,i)+\Delta ^{(1)} I(r,i)+\Delta ^{(1)} I(r,i+1)+ ...+\Delta ^{(1)} I(r,i+k-1)=
$
\vskip0.2cm
$
I(r,i+1)+\Delta ^{(1)} I(r,i+1)+ ...+\Delta ^{(1)} I(r,i+k-1)=....=I(r,i+k),
$
\vskip0.1cm
the last equality stands iterating the procedure.
\end{proof}

Moreover, in general, it is possible to write
\vskip0.2cm
$
\Delta ^{(1)} I(r,i)=I(r,i+1)-I(r,i)
$
\vskip0.2cm
$
\Delta ^{(2)} I(r,i)=[I(r,i+2)-I(r,i)]-[2\Delta ^{(1)}I(r,i)]
$
\vskip0.2cm
$
\Delta ^{(3)} I(r,i)=[I(r,i+3)-I(r,i)]-[3\Delta ^{(1)}I(r,i)+3\Delta ^{(2)}I(r,i)]
$
\vskip0.2cm
$
...
$
\begin{equation*}
\begin{split}
&\Delta ^{(k)} I(r,i)=[I(r,i+k)-I(r.i)]-[a_{k,1}\Delta ^{(1)}I(r,i)+...+\\
&+a_{k,j}\Delta ^{(j)}I(r,i) +...+a_{k,k-1}\Delta ^{(k-1)}I(r,i)]
\end{split}
\end{equation*}

\begin{figure} 
\begin{center} 
\includegraphics[scale=0.3]{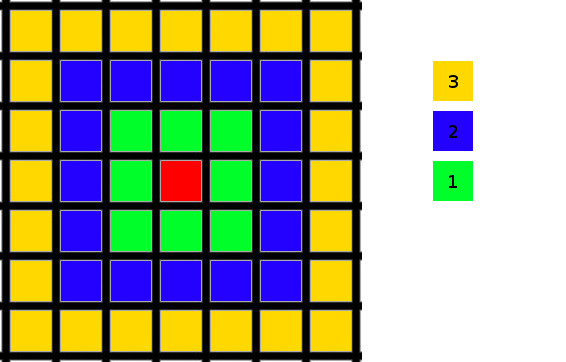}
\caption{\footnotesize An example of points involved in the calculation of finite differences: the red point of coordinates $(i,j)$ is the point to reconstruct; the green points identify the values needed for the calculation of the finite differences of order 1 respect to $(i,j)$; the blue points, together with the green points, identify the values needed for the calculation of the finite differences of order 2 respect to $(i,j)$; the yellow points, together with the blue and green points, identify the values needed for the calculation of the finite differences of order 3 respect to $(i,j)$.}
\label{fig:finiteDiff}
\end{center} 
\end{figure}

\vskip0.2cm
\hskip-0.5cm with the coefficients $\displaystyle a_{k,j}=\binom{k}{j}$, as they come form the Pascal's triangle formulation.
Given a direction, the above expression allows the calculation of the finite difference of order $k$ in $(r,i)$ as a linear combination of pointwise differences (i.e., $I(r,i+1)-I(r,i)$, $I(r,i+2)-I(r,i)$,...).
In the two dimensional case, this results in the combination of differences between the central point and a square shaped ring, depending on the maximum finite difference order chosen (see Fig. \ref{fig:finiteDiff}). In particular, this generally requires some interpolation or quasi interpolation techniques to provide values along those directions that are not explicitly available.

\begin{figure}
\begin{center}
\includegraphics[scale=0.3]{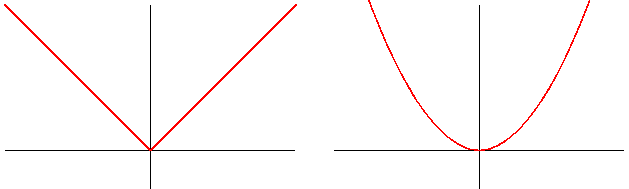}
\end{center}
\caption{\footnotesize A $C^0$ function on the left and a $C^k$ function on the right: a human observer can easily perceive the continuity of the left function but can not determine $k$.}
\label{fig:smoothness}
\end{figure}

	Calculating high order finite differences, in two dimensional setting, is time consuming. Moreover, recalling that the quality of the reconstruction is evaluated by simply observation, we stopped to the first order finite differences. In addition, to justify this choice also from a perceptive point of view, it is quite evident that a human observer can easily determine the class of continuity of a function only until the first order (see Fig. \ref{fig:smoothness}). The same reasons have been used to justify the definition (\ref{eq:DM}).

In the application it happens to have different candidate values with the same $E_T$: when this is the case, we choose their mean value as the best match. 
In Algorithm \ref{alg} we show the pseudo-code of the whole method theoretically described in the previous sections.

\begin{algorithm}
\caption{Inpaint by Finite Differences}
\label{alg}
\begin{algorithmic}[H]
\scriptsize
\Procedure{inpaint}{I,M,$T_S$,$\chi_S$,n}
\\
\Comment{I=Image}
\Comment{M=To-Be-Inpainted area}
\Comment{$T_S$=Training Set}
\Comment{$\chi_S$=support of Chi}
\Comment{n= order of finite differences}
\\
\State P=M
\\
\State $E_T \gets EnergyCalculation(I,M,T_S,\chi_S,n)$
\\
\Comment{first calculation of $E_T$}
\\
$loop:$		 \For{$(ii,jj) \in M$}
			\State $P\gets updatePriority(I,M,T_S,n,E_T,P)$
			\\		
			\Comment{return the priority for each point to inpaint}
			\\
			\State $I(find(P==max(P)) \gets T_S(bestFit(find(P==max(P)))$
			\\
			\Comment{inpaint the point with highest priority}
			\\
			\State $E_T \gets EnergyCalculation(I,M,T_S,\chi_S,n)$
			\\
			\Comment{update $E_T$ after the inpainting}
		\EndFor
$goto$ $loop:$
\EndProcedure

\\
\Procedure{$E_T \gets $EnergyCalculation}{I,M,$T_S$,$\chi_S$,n}
\\
\Comment{I=Image}
\Comment{M=To-Be-Inpainted area}
\Comment{$T_S$=Training Set}
\Comment{$\chi_S$=support of Chi}
\Comment{n= order of finite differences}
\\
\State bestFit=I
\\
	    \For{$(ii,jj) \in M$}
		\State $E_{T_{min}}=256\cdot ones(size(M))$
\\
	\For{$(i,j) \in T_S$}
			\State $E_C=sum(abs(I(ii \in \chi_S, jj \in \chi_S)-T_S(i \in \chi_S, j \in \chi_S)))$
			\State $E_{C_{norm}}=Norm(E_C) $ 
			\\
			\Comment{normalization of $E_C$ such that $ 0\leq E_C\leq 1$}
			\State $E_S=0$
			\For{d=1:n} 
				\For {k=1:8*k} 
					\State $E_S=E_S+sum(abs(\Delta(I(ii \in \chi_S, jj \in \chi_S),k,d)-\Delta(T_S(i\in \chi_S, j\ in \chi_S),k,d)))$
				\EndFor
			\EndFor
		
			\State $E_{S_{norm}}=Norm(E_S)$ 
			\\
			\Comment{normalization of $E_S$ such that $ 0\leq E_S\leq 1$}
			\State $E_{T_{*}}=E_{C_{norm}}+E_{S_{norm}}$
			\If{$ E_T <  E_{T_{min}}$}
				\State $bestFit(ii,jj) \gets T_S(i,j)$
				\State $E_T(ii,jj) \gets E_{T_{*}}$
			\EndIf
		\EndFor
	\EndFor
\EndProcedure
\\
\\
\Procedure{$P \gets $updatePriority}{I,M,$T_S$,$\chi_S$,n,$E_T$,P}
\\
\Comment{I=Image}
\Comment{M=To-Be-Inpainted area}
\Comment{$T_S$=Training Set}
\Comment{$\chi_S$=support of Chi}
\Comment{n= order of finite differences}
\Comment{$E_T$= Inpainting Energy}
\Comment{$P$= Priority}
\\
	\For{$(ii,jj) \in M$}
		\State $C(ii,jj) \gets calculateC(I,M,\chi_S)$
		\State $D(ii,jj,\theta)) \gets calculateD(I,M,\chi_S,n)$
		\State $P(ii,jj) \gets (C(ii,jj)+\max_{\theta}D(ii,jj,\theta))\cdot E_T(ii,jj)$
		\\
		\Comment{$C$ is the trustability, $\max_{\theta}D$ is the maximum of the finite differences; their calculation details are omitted for brevity}
	\EndFor
\EndProcedure
\end{algorithmic}
\end{algorithm}

The color images have been managed operating separately on the RGB channels and joined at the end of the procedure. No blending of any sort (e.g. Poisson blending) has been used to adjust the final results.

\section{Training set} \label{Training set}

A pivotal role, in the inpainting procedure, is played by the training set $T_S$. In particular, as the size of $T_S$ decreases, as the quality of the reconstructions degrades, due to the lower number of available patches. On the other hand, having a too big $T_S$ can result in wrong inpainted areas, especially in those points far enough from $\partial \Omega$. 

	Another drawback of a too extended $T_S$ is the computational time.

	Furthermore, the impressive results achieved in \cite{UVL} using neural networks prove that the information provided in images are redundant, such that even a limited subset is enough to inpaint correctly the missing areas.

\begin{figure}
\begin{center}

\hskip2.8cm
\includegraphics[scale=1]{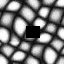}
\hskip0.2cm
\includegraphics[scale=1]{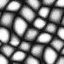}
\vskip0.2cm
\includegraphics[scale=1]{starting2.png}
\hskip0.5cm
\includegraphics[scale=1]{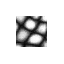}
\hskip0.2cm
\includegraphics[scale=1]{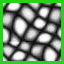}
\vskip0.2cm

\hskip2.9cm
\includegraphics[scale=1]{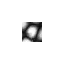}
\hskip0.2cm
\includegraphics[scale=1]{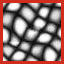}

\end{center}
\caption{\footnotesize Three examples of $T_S$ to inpaint the same $\Omega$ of size $14 \times 12$ (in black on the first column): on the left the image to inpaint; in the center the $T_S$ used to inpaint; on the right the achieved results. In this specific case the degradation, reducing $T_S$, appears evident only in the third case (highlighted in red, $2L+1=3$, $l=1$). The most pleasant reconstruction is the one highlighted in green, corresponding to a reduced $T_S$.}
\label{fig:reducingTS} 
\end{figure}

	Some results, with a reduced size of $T_S$, are shown in Fig. \ref{fig:reducingTS}: the lower extension of $T_S$ degrades the quality of the reconstructions only in the last case (third row, third column, highlighted in red). In the same figure, the best reconstruction appears to be the second one (second column, third row, highlighted in green). This result is counterintuitive because one would expect that more extended is $T_S$, better are the results, being higher the number of combinations among which operating the right choice. An explanation of this behavior is related to the possible presence of noise or to fluctuations of the light conditions, both altering the value of $E_T$ and introducing errors that propagate in $\Omega$ as the inpainting proceeds. Once these errors have been introduced, they can be compensated only by other random complementary fluctuations: when this does not happen the error is going to be amplified. To avoid this problem, different well known methods in the literature use some kind of smoothing (e.g., Gaussian smoothing in \cite{CPT}). 

\begin{figure}
\begin{center}

\hskip2.5cm
\includegraphics[scale=0.2]{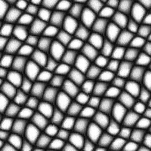}
\hskip0.cm
\includegraphics[scale=0.2]{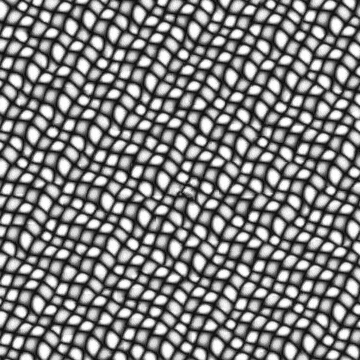}
\hskip0.cm
\includegraphics[scale=0.8]{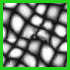}
\vskip0.2cm
\includegraphics[scale=0.2]{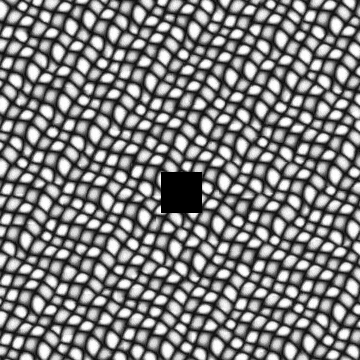}
\hskip0.5cm
\includegraphics[scale=0.2]{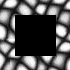}
\hskip0.cm
\includegraphics[scale=0.2]{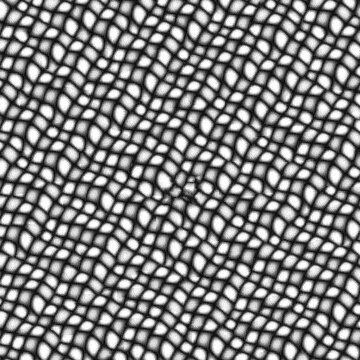}
\hskip0.cm
\includegraphics[scale=0.8]{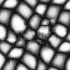}
\vskip0.2cm

\hskip3.5cm
\includegraphics[scale=0.2]{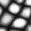}
\hskip0.cm
\includegraphics[scale=0.2]{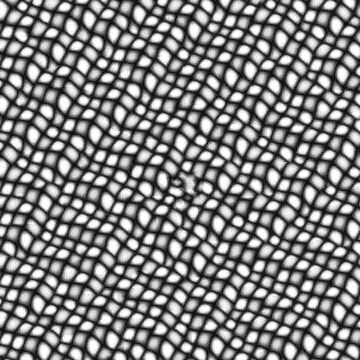}
\hskip0.cm
\includegraphics[scale=0.8]{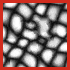}
\end{center}
\caption{\footnotesize Three examples of $T_S$ to inpaint the same $\Omega$ of size $41 \times 41$ pixels (in black on the first column): on the left the image to inpaint; in the second column the $T_S$ used to inpaint; in the third column the achieved results; on the right a zoomed detail of the results. In this specific case the degradation, reducing $T_S$, appears evident only in the third case (highlighted in red). The best reconstruction is highlighted in green and is achieved using the bigger $T_S$ ($2L+1=3$, $l=1$). In this case the size of $\supp \chi$ is not big enough to inpaint correctly. In Fig. \ref{fig:reducingTSbig} the correct result using $2L+1=5$, $l=1$.}
\label{fig:reducingTSbig} 
\end{figure}

	Another test, performed in case of $\Omega$ of bigger size, respect to Fig. \ref{fig:reducingTS}, is available in Fig. \ref{fig:reducingTSbig}: here the quality of the reconstructions increases with the size of $T_S$, but none of them is correct. In this example the problem is represented by the too small size of $\supp \chi$, more than by $T_S$ itself (see Fig. \ref{fig:reducingTSbig} for the correct reconstruction). 

	Moreover, advances in the field of texture synthesis (\cite{EFROS2}) have been based on the hypothesis that natural images are redundant, and self similar, such that a reduced $T_S$ appears to be an effective policy to correctly inpaint. In other works, it is stated that the inpainting process proceeds by continuation of the structure of the area surrounding $\Omega$ into it, such that contour lines are drawn via the prolongation of those arriving at $\partial \Omega$ (\cite{BSCB}). Moreover, other techniques have based their effectiveness on metrics including the proximity between the point to inpaint and the match found in $T_S$, to penalize the ones at higher distance (see, e.g., \cite{DIGESU,KSY}). 

	To the light of the previous observations, we can reduce $T_S$ only to surrounding areas of $\Omega$. In Sec. \ref{Numerical Results} each result will be shown together with the $T_S$ used to inpaint it.

\begin{figure}
\begin{center}
\includegraphics[scale=0.5]{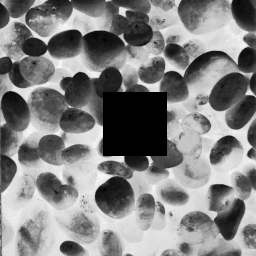}
\vskip0.1cm
\hskip1.05cm
\includegraphics[scale=0.5]{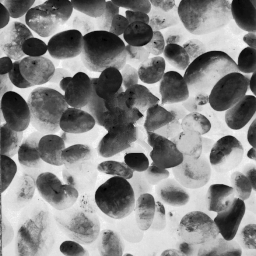}
\hskip0.1cm
\includegraphics[scale=0.5]{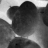}
\vskip0.1cm
\end{center}
\caption{\footnotesize Inpainting of a stone structure. From top: the image to be inpainted (top); the inpainting result with $2L+1=3$, $l=1$ (down left); the small $T_S$ used to inpaint (down right). We point out that the reconstruction, evidently not perfect, has been achieved using the only, randomly picked, small patch in the down right (basically one single hole of the whole structure).}
\label{fig:TS3} 
\end{figure}

\begin{figure}
\begin{center}
\includegraphics[scale=1]{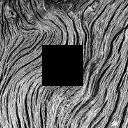}
\vskip0.1cm
\hskip1.3cm
\includegraphics[scale=1]{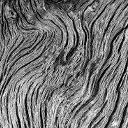}
\hskip0.1cm
\includegraphics[scale=1]{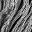}
\vskip0.1cm
\end{center}
\caption{\footnotesize Inpainting of a wood patch. From top: the image to be inpainted (top); the inpainting result with $2L+1=3$, $l=1$ (down left); the small $T_S$ used to inpaint (down right). We point out that the reconstruction, evidently not perfect, has been achieved using the only, randomly picked, small patch in the down right.}
\label{fig:wood} 
\end{figure}

\begin{figure}
\begin{center}
\includegraphics[scale=0.8]{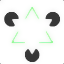}
\hskip0.3cm
\includegraphics[scale=0.8]{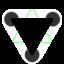}
\hskip0.3cm
\includegraphics[scale=0.8]{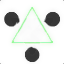}

\end{center}
\caption{\footnotesize Inpainting of the Kanizsa triangle. The starting image (left); the $T_S$ (center); the inpainting result, achieved using $2L+1=3$, $l=1$ (right). The ticker borders in the reconstruction are due to the mean operation used in case of multiple candidate values.}
\label{fig:KNSZ} 
\end{figure}

More examples, using a reduced $T_S$, are available from Fig.\ref{fig:TS3} to Fig. \ref{fig:KNSZ}.

\section{Numerical Results} \label{Numerical Results}

In this section we show numerical results achieved using $E_T$. No limitations are imposed on the topology of $\Omega$.
\vskip0.3cm

\begin{figure}
\begin{center}
\includegraphics[scale=0.1]{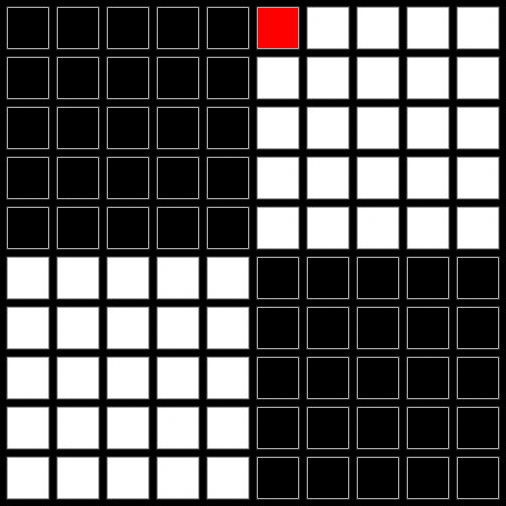}
\end{center}
\caption{\footnotesize In red, the point to inpaint in the periodic chessboard-like pattern. Each square of the chessboard consists of $5\times 5$ pixels.}
\label{fig:chess} 
\end{figure}

	The first example inpaints one single point of coordinates $(i,j)$ in a periodic chessboard-like pattern, as the red one in Fig. \ref{fig:chess}. 
As result of the reconstruction, we expect a white value to fill the missing point. To provide an example, we take $2L+1=3$. According to what stated in the theoretical section, we assume $(i,j)$ to be possibly located in whatever position of $\supp \chi$ and not necessarily in its center. We name the configurations arising from this choice from 3.1 to 3.9 and we group them in the set of patterns $S_C=\{ 3.1,3.2,3.3,3.4,3.5,3.6,3.7,3.8,3.9\}$. Each element of $S_C$ is marked by a couple of integers separated by a dot: the first number gives the size of one side of $\supp \chi$, the second one specifies the pattern position in the set. In fact, $S_C$ individuates the equivalence classes mentioned in Sec. \ref{Neighborhood}. 

\begin{figure}
\begin{center}
\includegraphics[scale=0.2]{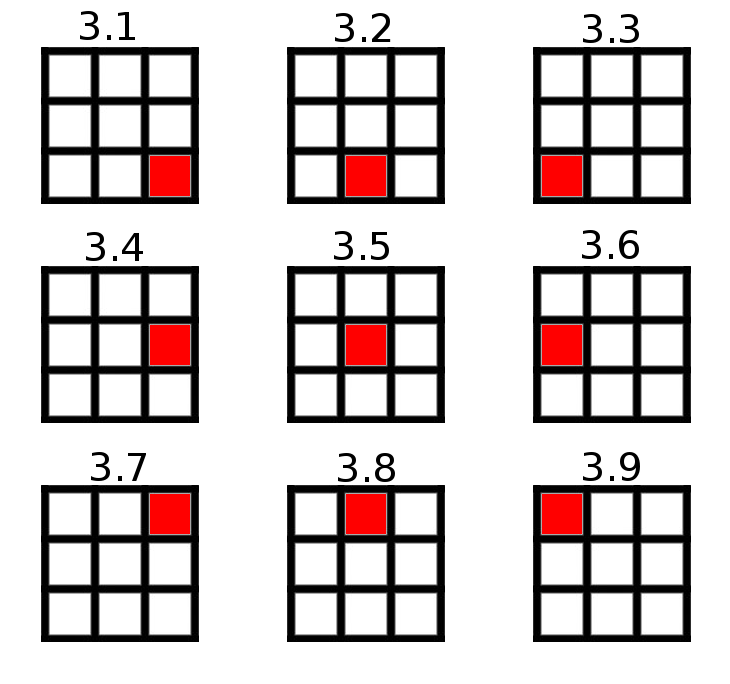}
\hskip0.9cm
\includegraphics[scale=0.2]{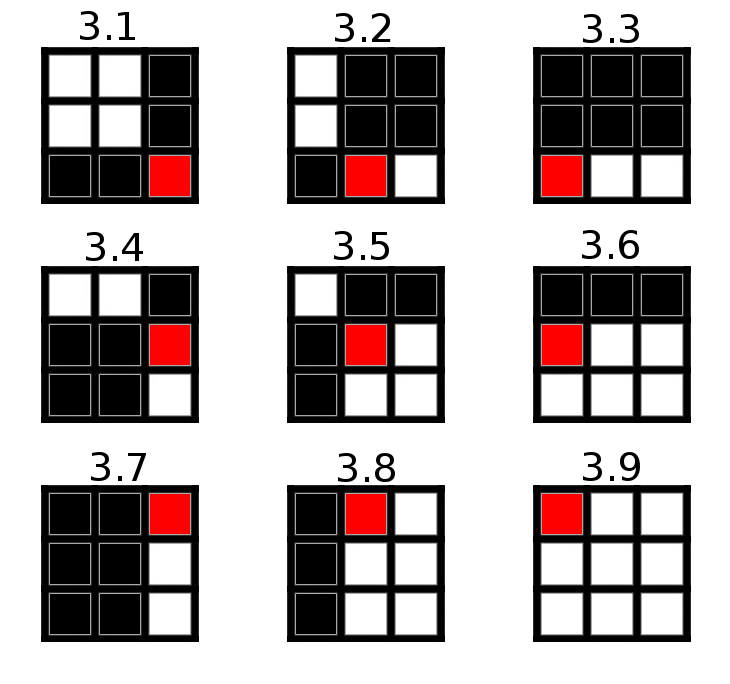}
\end{center}
\caption{\footnotesize On the left: all the possible position of the missing point (in red) in $\supp \chi$. On the right: each $3\times 3$ sized pattern compatible with the missing point of Fig. \ref{fig:chess}.}
\label{fig:patterns} 
\end{figure}

	The number and elements of this set are shown in Fig. \ref{fig:chess}, for $k=1$.  Being the red point surrounded by known values, each configuration of the set $S_C$ is causal  (refer to \cite{DEM} for a further definition of causality applyed to inpainting). A different choice of configuration of $S_C$  will work differently but we expect to achieve the same final result (i.e., a white point). The patches compatible with the configurations $S_C$ and giving a correct solution, taken from the chessboard of Fig. \ref{fig:chess} (representing our training set $T_S$) \footnote{One more time, we use the expression \textit{training set} borrowing it from the Neural Networks (NN) literature, meaning that the missing points in the inpainting process are deduced from the  given data contained in $T_s$ that plays, this way, similarly to a NN training set.}, are shown in the right part of Fig. \ref{fig:patterns}.

\begin{figure}
\begin{center}
\includegraphics[scale=0.1]{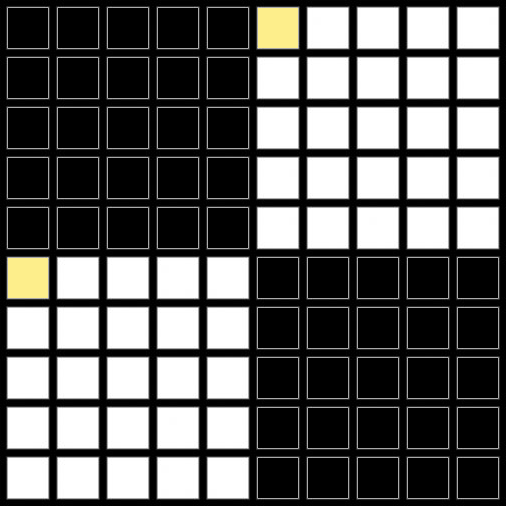}
\includegraphics[scale=0.1]{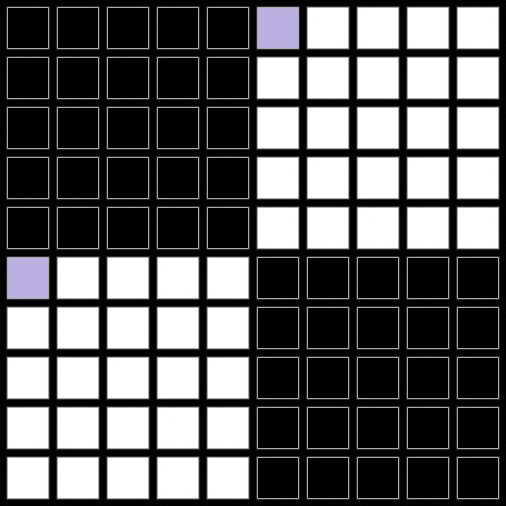}
\includegraphics[scale=0.1]{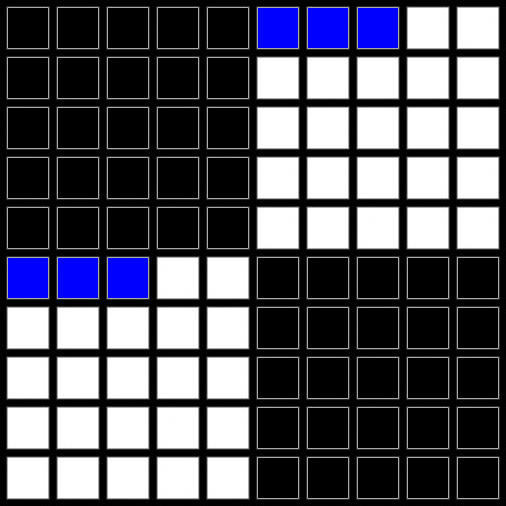}
\vskip0.1cm
\includegraphics[scale=0.1]{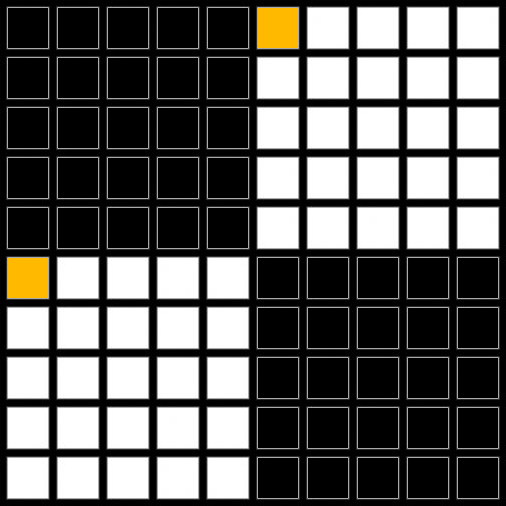}
\includegraphics[scale=0.1]{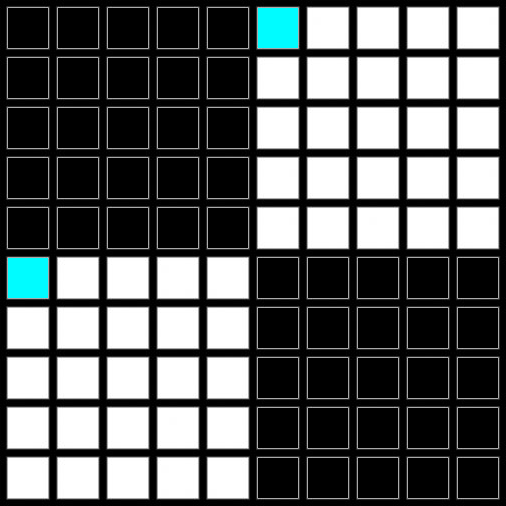}
\includegraphics[scale=0.1]{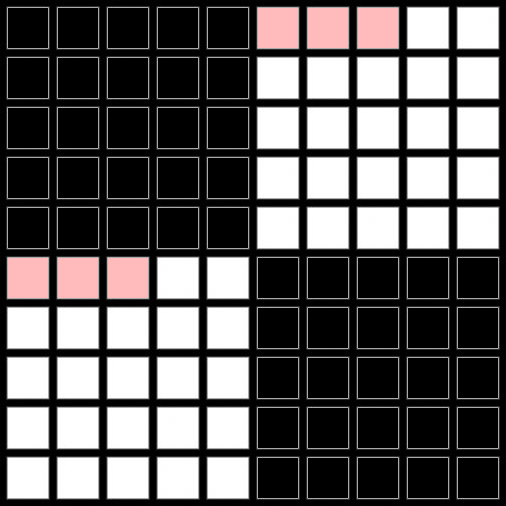}
\vskip0.1cm
\includegraphics[scale=0.1]{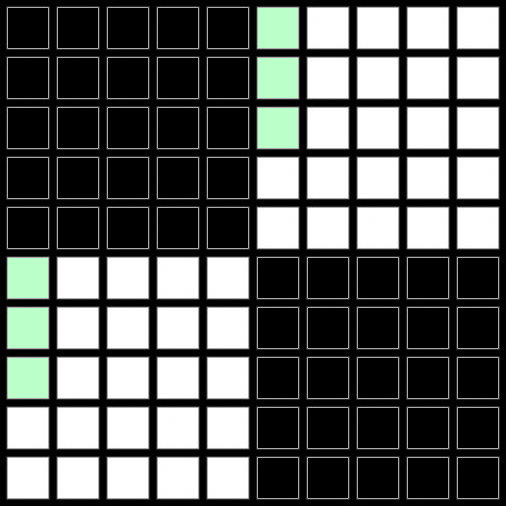}
\includegraphics[scale=0.1]{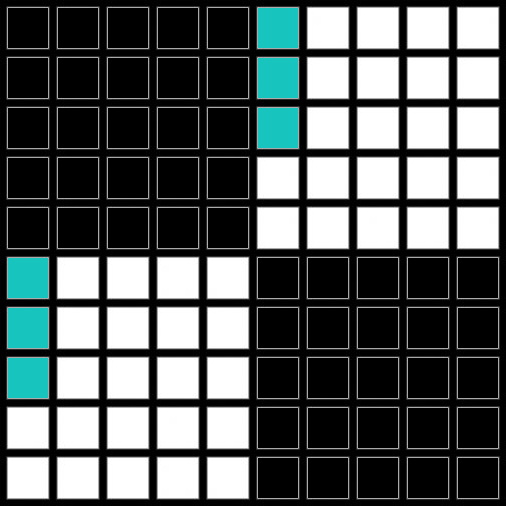}
\includegraphics[scale=0.1]{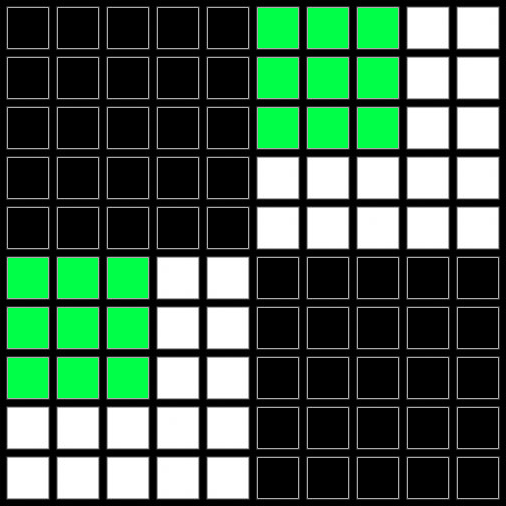}
\end{center}
\caption{\footnotesize From top to down, left to right: the candidate points for each different pattern (from 3.1 to 3.9). The squares with the same color in the 9 chessboards constitute the sets $\mathcal{S}_{3.1}$, $\mathcal{S}_{3.2}$, ... ,$\mathcal{S}_{3.9}$.}
\label{fig:solutions} 
\end{figure}

To each configuration of $S_C$, is associated a set of points compatible with the reconstruction, given $T_S$: for each element of $S_C$ they are shown using different colors in Fig. \ref{fig:solutions}: we name each set with $\mathcal{S}_{3.1}$, $\mathcal{S}_{3.2}$, ... ,$\mathcal{S}_{3.9}$.

The exact solution, i.e., the one taking into account as the content (black or white color), as the structure (i.e., the fact that the missing point is on a corner) comes from sets $\mathcal{S}_{3.1},\mathcal{S}_{3.2},\mathcal{S}_{3.4},\mathcal{S}_{3.5}$ only. The other sets as well correctly identify the white value as the correct one but the result is achieved not taking into account the particular position of the red point in the chessboard. For this reason, in this particular case, being perfect the correspondence between neighborhoods, $E_T=E_{C_{T}}$ is enough to achieve a visually correct result. In more complex and general cases this is not true.  Moreover, the size of $\supp \chi$ is not relevant for the correctness of the reconstruction (i.e., the minimum possible size $2L+1=3$ is enough). On the other hand, when there is not a perfect match in $T_S$, the role played by $E_{S_{T}}$ increases in its importance, such that considering $E_T=E_{C_{T}}+E_{S_{T}}$ is {crucial}.

\begin{figure}
\begin{center}
\includegraphics[scale=1]{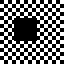}
\includegraphics[scale=1]{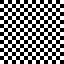}
\end{center}
\caption{\footnotesize Inpainting of the missing part $\Omega$ in a chessboard, using $2L+1=3$, $l=1$.}
\label{fig:inpChess}
\end{figure}

	An example of the inpainting of a chessbord with our method is provided in Fig. \ref{fig:inpChess}. To support this statement other numerical cases are considered.
\vskip0.3cm

\begin{figure}
\begin{center}
\includegraphics[scale=0.3]{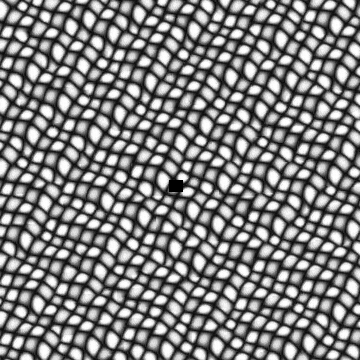}
\hskip0.1cm
\includegraphics[scale=1.69]{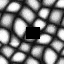}
\end{center}
\caption{\footnotesize The area to be inpainted (in black) in the context of the entire image (left) and in detail (right).}
\label{fig:numprob} 
\end{figure}

In a quasi periodic texture (de Bonet's sample number 161 \cite{DEB}), $\Omega$ has size of $12\times 14$ pixels (the black rectangle in Fig. \ref{fig:numprob}). 

\begin{figure}
\begin{center}
\includegraphics[scale=0.3]{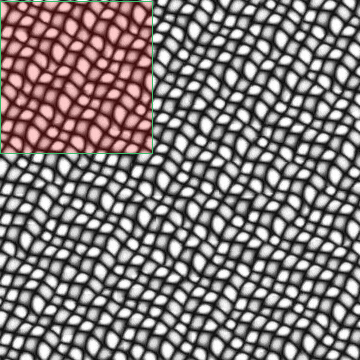}
\end{center}
\caption{\footnotesize On the top left corner, highlighted in red, the training set $T_S$ we used to inpaint the quasi periodic texture.}
\label{fig:trainingset} 
\end{figure}

To reduce the calculation time, we extract from the starting image an opportune random sample $T_S$ assumed to be representative of the whole texture (highlighted in red in Fig. \ref{fig:trainingset}).

\begin{figure}
\begin{center}
\includegraphics[scale=1.3]{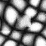} 
\hskip0.1cm
\includegraphics[scale=1.3]{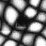} 
\end{center}

\caption{\footnotesize Inpainting of $\Omega$ (shown in Fig. \ref{fig:numprob}) using \cite{WEI}, for different values of $\supp \chi$ and different initial conditions. On the left: initial condition with random white noise; on the right:  initial condition equal to zero. Both images achieved using $2L+1=3$. This is an example of missing causality condition.}
\label{fig:ic} 
\end{figure}

The size of $\supp \chi$ influences the correctness of the final reconstruction but, to the best of our knowledge, there does not exist a consolidated methodology to estimate its optimal size.
Moreover, the quality of the reconstruction is connected with the visual inspection of the result by a human observer, such that different solutions can be considered qualitatively equivalent. The size of $\supp \chi$ plays a role more important than the initial conditions as discriminating for content as it is for structure: in Fig. \ref{fig:ic} inpainting results are shown as using a random white noise for the initial condition, as putting each point of $\Omega$ to a constant value (zero in this case). From the same figure comes that the quality of the final reconstruction is mostly determined by the size of $\supp \chi_{i,j}$ than by the initial condition (both the results in figure are not acceptable). 

	To control the simulations and to have reproducible and comparable results, we chose to use the constant value zero as initial condition for $\Omega$ in our algorithm. The respect of the causality condition assures that the inpainting process is independent from the initial conditions.
\vskip0.3cm

\begin{figure}
\begin{center}
\includegraphics[scale=0.8]{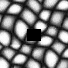}
\hskip0.1cm
\includegraphics[scale=0.8]{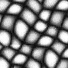}
\hskip0.1cm
\includegraphics[scale=0.8]{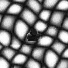}
\end{center}
\caption{\footnotesize From left to right: initial $\Omega$ of size $12 \times 14$ pixels (left); inpainting result using out method ($2L+1=3$, $l=1$) (center); inpainting result using method \cite{WEI}  with $2L+1=11$ (right). The better performance deriving from the introduction of $E_{S_{T}}$ and the causality of the new proposed method is visually evident. The image in the center is achieved at the first iteration and does not change meaningfully in the subsequent ones, highlighting a fast convergence too.}
\label{fig:dorder1} 
\end{figure}

Using the method proposed in \cite{WEI}, as the size of the neighborhood increases, as the reconstruction quality improves (minimum size for a correct inpainting is $2L+1=13$).
Using our formalization with $E_T=E_{C_{T}}+E_{S_{T}}$, in place of $E_T=E_{C_{T}}$, even the case $2L+1=3$ results in a correct reconstruction (see Fig. \ref{fig:dorder1}). The contribution of the finite differences, as well as the respect of the causality condition, decrease the uncertainty in the choice of the correct value, determining good results even using the minimum size for $\supp \chi$.

\begin{figure}
\begin{center}
\includegraphics[scale=0.5]{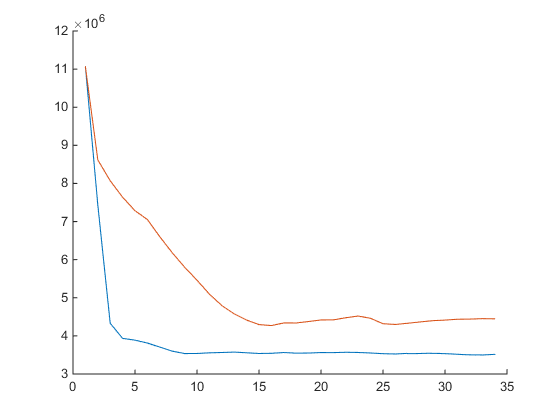}
\end{center}
\caption{\footnotesize Figure shows $E_T$ trends, calculated at the end of each scanning of the whole $\Omega$: in red $E_T=E_{C_{T}}$, in  blue $E_T=E_{C_{T}}+E_{S_{T}}$. Finite differences contributions determine a faster convergence to the steady state.}
\label{fig:trueEN11} 
\end{figure}

In \cite{DEM} they formulated the expression of the  inpainting energy and showed that it has a decreasing trend. We do the same, achieving that our new version $E_T=E_{C_{T}}+E_{S_{T}}$ not only has a decreasing trend but it reaches the minimum value faster (see Fig. \ref{fig:trueEN11}). The trend shown in Fig. \ref{fig:trueEN11} is qualitatively representative of all the performed tests.  
\begin{remark}
 The value of $E_T$ is meaningful when it is calculated at the end of each scan and if the causality condition is respected. Otherwise, in the inpanting algorithm, a potentially dangerous (for stability) positive feedback is introduced. During the first scan the causality property addresses the algorithm to the direction of that minimum characterized by the lowest uncertainty. Subsequent scans refine the calculation.
\end{remark}

\begin{figure}
\begin{center}
\includegraphics[scale=0.7]{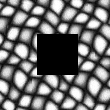}
\hskip0.1cm
\includegraphics[scale=0.7]{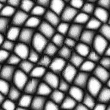}
\hskip0.1cm
\includegraphics[scale=0.7]{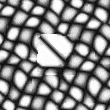}
\end{center}
\caption{\footnotesize From left to right: initial $\Omega$ (in black), size $41 \times 41$ pixels (left); inpainting result using our method ($2L+1=5$, $l=1$) (center); inpainting result using the method  proposed in \cite{WEI}, with $2L+1=11$. The better performance of our algorithm is evident. The big differences are localized "far enough" from $\partial \Omega$, i.e., in those points where the uncertainty increases.}
\label{fig:41x41} 
\end{figure}

 The introduction of $E_{S_{T}}$ allows to carry more structural information in the similarity computation.

\begin{figure}[!htb]
\begin{center}
\includegraphics[scale=0.45]{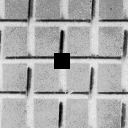}
\hskip0.01cm
\includegraphics[scale=0.45]{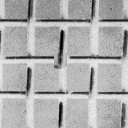}
\hskip0.01cm
\includegraphics[scale=0.45]{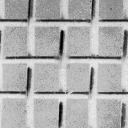}
\hskip0.01cm
\includegraphics[scale=0.45]{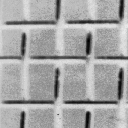}
\end{center}
\caption{\footnotesize  Inpainting of texture D1 taken from Brodatz database. From left to right: image to be inpainted ($\Omega$ in black, size $16 \times 16$); inpainting by mean of $E_T=E_{C_{T}}$ with $2L+1=23$ (second column) and $E_T=E_{C_{T}}+E_{S_{T}}$ ($2L+1=3$, $l=1$) (third column); training set $T_S$ (size $128 \times 128$) in the fourth column. The case using the first order finite difference reconstructs better because it does not introduce structures that are not present in $T_S$. Moreover the minimum value of $L$ to inpaint correctly is 23 when $E_T=E_{C_{T}}$, against $L=3$ when introducing $E_{S_{T}}$. Finally, according to the fact that $T_S$ is more blurred than the image to fill, the inpainting results appear slightly more blurred than the expected ones.}
\label{fig:D1} 
\end{figure}

\begin{figure}[!htb]
\begin{center}
\includegraphics[scale=0.5]{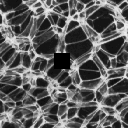}
\hskip0.01cm
\includegraphics[scale=0.5]{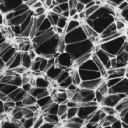}
\hskip0.01cm
\includegraphics[scale=0.5]{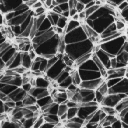}
\hskip0.10cm
\includegraphics[scale=0.5]{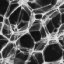}
\end{center}
\caption{\footnotesize Inpainting of texture D111 taken from Brodatz database. From left to right: image to be inpainted ($\Omega$ in black, size $16 \times 16$); inpainting by mean of $E_T=E_{C_{T}}$ with $2L+1=27$ (second column); inpainting by mean of $E_T=E_{C_{T}}+E_{S_{T}}$ with $2L+1=3$, $l=1$ (third column); $T_S$ (size $64 \times 64$) in the fourth column. The case using the first order finite difference reconstructs visually better. Note, again, the different value of $L$.}
\label{fig:D111} 
\end{figure}

	Other examples of reconstructions for quasi periodic textures are provided in Figs. from \ref{fig:41x41} to \ref{fig:D111}. 

\begin{figure}[!htb]
\begin{center}
\includegraphics[scale=0.8]{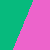}
\hskip0.1cm
\includegraphics[scale=0.8]{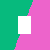}
\hskip0.1cm
\includegraphics[scale=0.8]{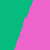}
\hskip0.1cm
\includegraphics[scale=0.8]{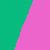}
\hskip0.1cm
\includegraphics[scale=0.8]{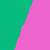}
\end{center}
\caption{\footnotesize From left to right: original image, image to be inpainted ($\Omega$ in white), reconstruction by \cite{CPT}, reconstruction by Matlab \textit{inpaintExemplar} function, reconstruction using our method ($2L+1=3 \times 3$, $l=1$). All the methods guarantee the connection of the isophote along a straight line, according to the connectivity principle.}
\label{fig:p} 
\end{figure}

\begin{figure}[!htb]
\begin{center}
\includegraphics[scale=1.35]{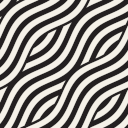}
\hskip0.1cm
\includegraphics[scale=1.35]{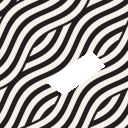}
\hskip0.1cm
\includegraphics[scale=0.325]{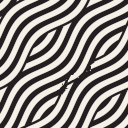}
\hskip0.1cm
\includegraphics[scale=0.325]{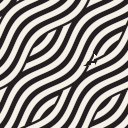}
\hskip0.1cm
\includegraphics[scale=0.325]{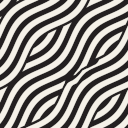}

\vskip0.5cm

\includegraphics[scale=1.265]{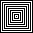}
\hskip0.1cm
\includegraphics[scale=1.265]{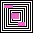}
\hskip0.1cm
\includegraphics[scale=1.265]{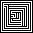}
\hskip0.1cm
\includegraphics[scale=1.265]{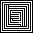}
\hskip0.1cm
\includegraphics[scale=1.265]{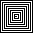}
\end{center}
\caption{\footnotesize From left to right: original image, image to be inpainted ($\Omega$ in white (top) and pink (down)), reconstruction by \cite{CPT}, reconstruction by Matlab \textit{inpaintExemplar} function, reconstruction using our method ($2L+1=3$, $l=1$). Analogous results are achieved changing the size of $\supp \chi$ for the comparison methods. Considering the known part of the image, the reconstructions using the proposed method exhibits more regularity respect with the other two algorithms.}
\label{fig:maze_zigzag} 
\end{figure}

\begin{figure}[!htb]
\begin{center}
\includegraphics[scale=0.6]{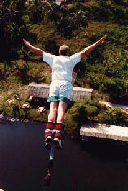}
\hskip0.1cm
\includegraphics[scale=0.6]{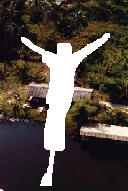}
\hskip0.1cm
\includegraphics[scale=0.6]{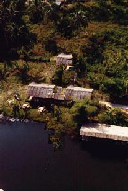}
\vskip0.1cm
\includegraphics[scale=0.6]{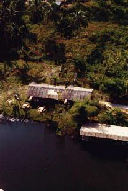}
\hskip0.1cm
\includegraphics[scale=0.6]{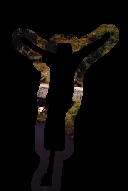}
\hskip0.1cm
\includegraphics[scale=0.6]{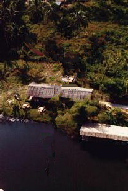}
\end{center}
\caption{\footnotesize Reconstructions from different inpainting methods. First row: original image (left); impage to inpaint with $\Omega$ in black (center); reconstruction by \cite{CPT} (right). Second row:  Matlab \textit{inpaintExemplar} reconstruction (left); $T_S$ used in our method (center); reconstruction by our method ($2L+1=3$, $l=1$) (right).}
\label{fig:input} 
\end{figure}

\begin{figure}[!htb]
\begin{center}
\includegraphics[scale=0.54]{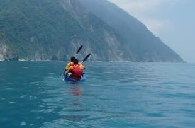}
\hskip0.1cm
\includegraphics[scale=0.54]{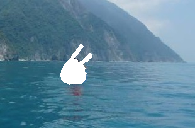}
\hskip0.1cm
\includegraphics[scale=0.405]{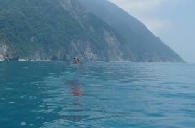}
\vskip0.1cm
\includegraphics[scale=0.405]{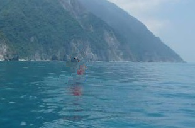}
\hskip0.1cm
\includegraphics[scale=0.54]{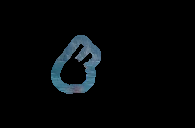}
\hskip0.1cm
\includegraphics[scale=0.405]{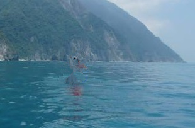}
\end{center}
\caption{\footnotesize Reconstructions from different inpainting methods. First row: original image (left); impage to inpaint with $\Omega$ in black (center); reconstruction by \cite{CPT} (right). Second row:  Matlab \textit{inpaintExemplar} reconstruction (left); $T_S$ used in our method (center); reconstruction by our method ($2L+1=3$, $l=1$) (right).}
\label{fig:3} 
\end{figure}

\begin{figure}[!htb]
\begin{center}
\includegraphics[scale=0.61]{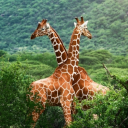}
\hskip0.1cm
\includegraphics[scale=0.61]{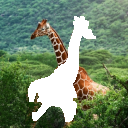}
\hskip0.1cm
\includegraphics[scale=0.61]{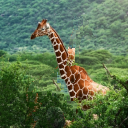}
\vskip0.1cm
\includegraphics[scale=0.61]{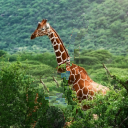}
\hskip0.1cm
\includegraphics[scale=0.61]{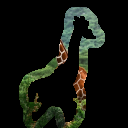}
\hskip0.1cm
\includegraphics[scale=0.61]{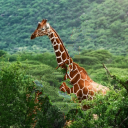}
\end{center}
\caption{\footnotesize Reconstructions from different inpainting methods. First row: original image (left); impage to inpaint with $\Omega$ in black (center); reconstruction by \cite{CPT} (right). Second row:  Matlab \textit{inpaintExemplar} reconstruction (left); $T_S$ used in our method (center); reconstruction by our method ($2L+1$, $l=1$) (right).}
\label{fig:giraffes} 
\end{figure}

\begin{figure}[!htb]
\begin{center}
\includegraphics[scale=0.6]{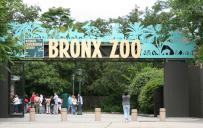}
\hskip0.1cm
\includegraphics[scale=0.6]{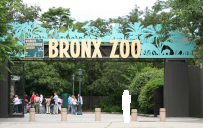}
\hskip0.1cm
\includegraphics[scale=0.6]{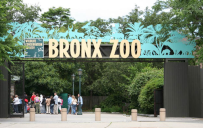}
\hskip0.1cm
\includegraphics[scale=0.6]{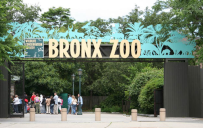}
\vskip0.1cm
\includegraphics[scale=0.6]{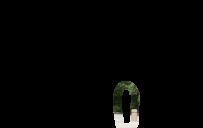}
\hskip0.1cm
\includegraphics[scale=0.6]{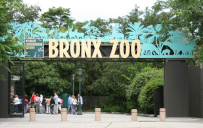}
\end{center}
\caption{\footnotesize Reconstructions from different inpainting methods. First row: original image (left); impage to inpaint with $\Omega$ in black (center); reconstruction by \cite{CPT} (right). Second row:  Matlab \textit{inpaintExemplar} reconstruction (left); $T_S$ used in our method (center); reconstruction by our method ($2L+1=3$, $l=1$) (right).}
\label{fig:zoo} 
\end{figure}

\begin{figure}[!htb]
\begin{center}
\includegraphics[scale=0.35]{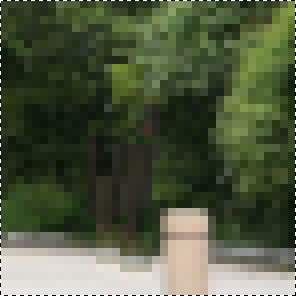}
\hskip0.1cm
\includegraphics[scale=0.35]{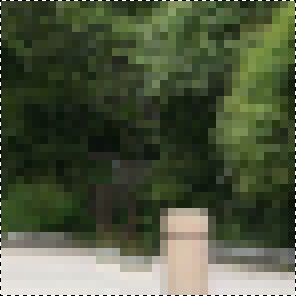}
\hskip0.1cm
\includegraphics[scale=0.35]{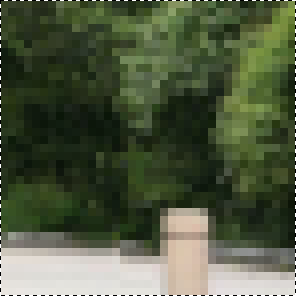}
\end{center}
\caption{\footnotesize  Details of the reconstructions shown in Fig. \ref{fig:zoo} (left side of the stem of the mushroom). From left to right: Matlab \textit{inpaintExemplar} (left); \cite{CPT} (center); our method (right). On one hand, at this level of zoom, a light shadow of the inpainted area appears in the first two images; on the other hand, the inpainted area is not detectable by a human observer looking at the reconstruction achieved by our method.}
\label{fig:detail_zoo} 
\end{figure}

\begin{figure}[!htb]
\begin{center}
\includegraphics[scale=0.6]{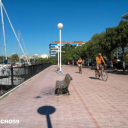}
\hskip0.1cm
\includegraphics[scale=0.6]{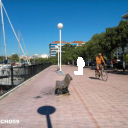}
\hskip0.1cm
\includegraphics[scale=0.6]{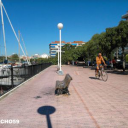}
\vskip0.1cm
\includegraphics[scale=0.6]{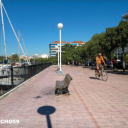}
\hskip0.1cm
\includegraphics[scale=0.6]{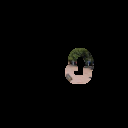}
\hskip0.1cm
\includegraphics[scale=0.6]{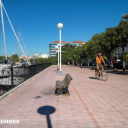}
\end{center}
\caption{\footnotesize Reconstructions from different inpainting methods. First row: original image (left); impage to inpaint with $\Omega$ in black (center); reconstruction by \cite{CPT} (right). Second row:  Matlab \textit{inpaintExemplar} reconstruction (left); $T_S$ used in our method (center); reconstruction by our method ($2L+1=3$, $l=1$) (right).}
\label{fig:walk} 
\end{figure}

\begin{figure}[!htb]
\begin{center}
\includegraphics[scale=0.2305]{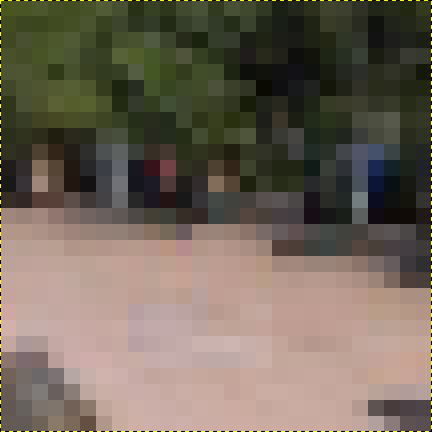}
\hskip0.1cm
\includegraphics[scale=0.2305]{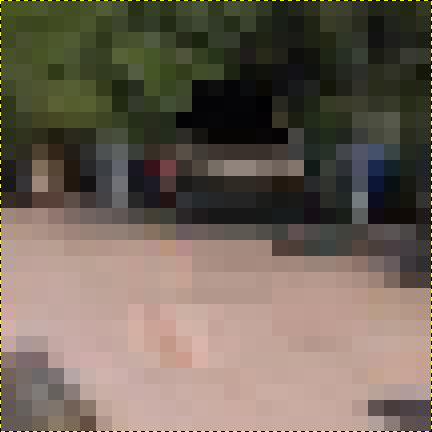}
\hskip0.1cm
\includegraphics[scale=0.2305]{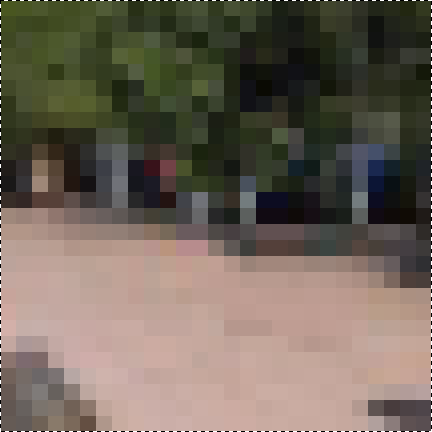}
\end{center}
\caption{\footnotesize Details of the reconstructions shown in Fig. \ref{fig:walk} (left side of the stem of the mushroom). From left to right: Matlab \textit{inpaintExemplar} (left); \cite{CPT} (center); our method (right). On one hand, spurious structures appear in the first two images, being detectable by a human observer, at this level of zoom; on the other hand, the reconstruction by our method looks completely natural.}
\label{fig:detail_walk} 
\end{figure}

\begin{figure}[!htb]
\begin{center}
\includegraphics[scale=2.5]{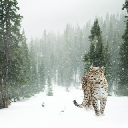}
\hskip0.1cm
\includegraphics[scale=2.5]{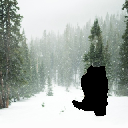}
\hskip0.1cm
\includegraphics[scale=0.6]{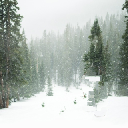}
\hskip0.1cm
\includegraphics[scale=0.6]{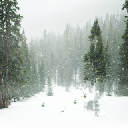}
\hskip0.1cm
\includegraphics[scale=2.49]{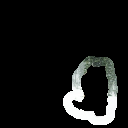}
\hskip0.1cm
\includegraphics[scale=0.6]{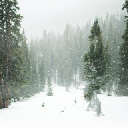}
\end{center}
\caption{\footnotesize Reconstructions from different inpainting methods. First row: original image (left); impage to inpaint with $\Omega$ in black (center); reconstruction by \cite{CPT} (right). Second row:  Matlab \textit{inpaintExemplar} reconstruction (left); $T_S$ used in our method (center); reconstruction by our method ($2L+1=3$, $l=1$) (right).}
\label{fig:leo} 
\end{figure}

\begin{figure}[!htb]
\begin{center}
\includegraphics[scale=0.6]{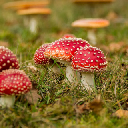}
\hskip0.1cm
\includegraphics[scale=0.6]{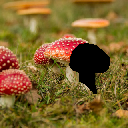}
\hskip0.1cm
\includegraphics[scale=0.6]{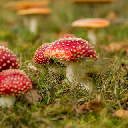}
\vskip0.1cm
\includegraphics[scale=0.6]{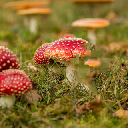}
\hskip0.1cm
\includegraphics[scale=0.6]{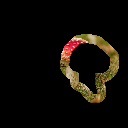}
\hskip0.1cm
\includegraphics[scale=0.6]{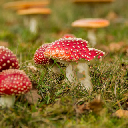}
\end{center}
\caption{\footnotesize Reconstructions from different inpainting methods. First row: original image (left); impage to inpaint with $\Omega$ in black (center); reconstruction by \cite{CPT} (right). Second row:  Matlab \textit{inpaintExemplar} reconstruction (left); $T_S$ used in our method (center); reconstruction by our method ($2L+1=3$, $l=1$) (right).}
\label{fig:mush} 
\end{figure}

\begin{figure}[!htb]
\begin{center}
\includegraphics[scale=0.27]{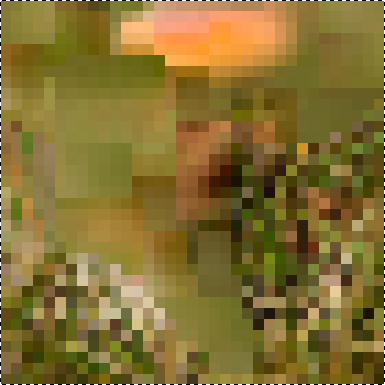}
\hskip0.1cm
\includegraphics[scale=0.27]{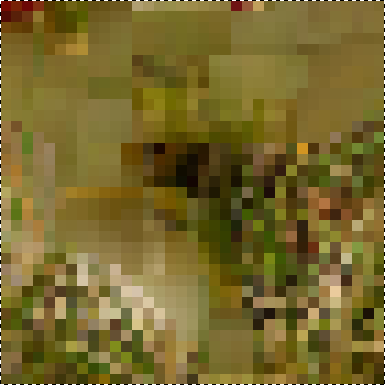}
\hskip0.1cm
\includegraphics[scale=0.27]{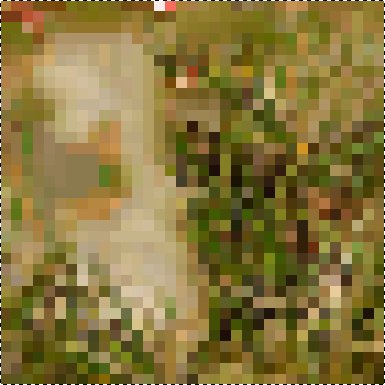}
\end{center}
\caption{\footnotesize Details of the reconstructions shown in Fig. \ref{fig:mush} (left side of the stem of the mushroom). From left to right: Matlab \textit{inpaintExemplar} (left); \cite{CPT} (center); our method (right). Our reconstruction is less blurred and preserves the thin structures of the grass. The interesting thing, in this example, is that our method attempts  to generate a new mushroom, in place of the deleted one.}
\label{fig:detail_mush} 
\end{figure}

	Moreover, the proposed method has been compared with results achieved by Matlab\copyright\space \textit{inpaintExemplar} function (\cite{CPT,LEG}), and with ones coming from an available implementation of \cite{CPT} (see Figs. from \ref{fig:p}  to \ref{fig:detail_mush}).
	
The last examples, from Fig. \ref{fig:snow} to Fig. \ref{fig:bridge}, provide a comparison with some other well known methods.

\begin{figure}[!h]

\end{figure}

\begin{figure}[!htb]
\begin{center}
    \begin{subfigure}[b]{0.235\textwidth}
        \includegraphics[width=\textwidth]{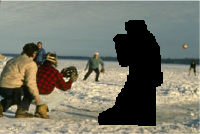}
   \caption{\footnotesize To be inpainted.}
  \label{snow:To be inpainted}
    \end{subfigure}
   \hskip0.1cm
    \begin{subfigure}[b]{0.235\textwidth}
        \includegraphics[width=\textwidth]{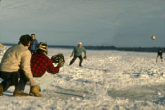}
   \caption{\footnotesize Inpainted by \cite{PGB}.}
  \label{snow:PGB}
    \end{subfigure}
\vskip0.2cm
    \begin{subfigure}[b]{0.235\textwidth}
        \includegraphics[width=\textwidth]{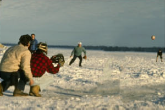}
   \caption{\footnotesize Inpainted by \cite{H}.}
  \label{snow:H}
    \end{subfigure}
   \hskip0.1cm
    \begin{subfigure}[b]{0.235\textwidth}
        \includegraphics[width=\textwidth]{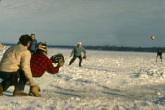}
   \caption{\footnotesize Inpainted by \cite{KT}.}
  \label{snow:PGB}
    \end{subfigure}
\vskip0.2cm
    \begin{subfigure}[b]{0.235\textwidth}
        \includegraphics[width=\textwidth]{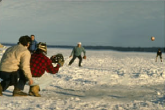}
   \caption{\footnotesize Inpainted by \cite{AFCS}  NL-means.}
  \label{snow:NLM}
    \end{subfigure}
   \hskip0.1cm
    \begin{subfigure}[b]{0.235\textwidth}
        \includegraphics[width=\textwidth]{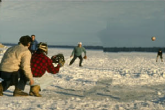}
   \caption{\footnotesize Inpainted by \cite{AFCS} NL-medians.}
  \label{snow:NLMd}
    \end{subfigure}
\vskip0.2cm
    \begin{subfigure}[b]{0.235\textwidth}
        \includegraphics[width=\textwidth]{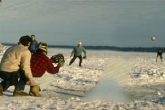}
   \caption{\footnotesize Inpainted by \cite{AFCS}  NL-Poisson.}
  \label{snow:NLP}
    \end{subfigure}
   \hskip0.1cm
    \begin{subfigure}[b]{0.235\textwidth}
        \includegraphics[width=\textwidth]{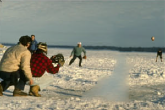}
   \caption{\footnotesize Inpainted by \cite{AFCS} NL-GM.}
  \label{snow:NLGM}
    \end{subfigure}
\vskip0.2cm
    \begin{subfigure}[b]{0.235\textwidth}
        \includegraphics[width=\textwidth]{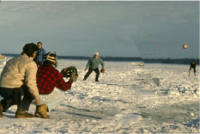}
   \caption{\footnotesize Inpainted by our method ($2L+1=3$, $l=1$).}
  \label{snow:our}
    \end{subfigure}

\end{center}
\caption{\footnotesize Inpainting by different well known methods in the literature. Images from (c) to (h) taken from \cite{AFCS}, with $2L+1>3$. In our method $2L+1=3$.}
\label{fig:snow} 
\end{figure}

\begin{figure}[!htb]
\begin{center}
    \begin{subfigure}[b]{0.235\textwidth}
        \includegraphics[width=\textwidth]{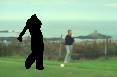}
   \caption{\footnotesize To be inpainted.}
  \label{golf:To be inpainted}
    \end{subfigure}
   \hskip0.1cm
    \begin{subfigure}[b]{0.235\textwidth}
        \includegraphics[width=\textwidth]{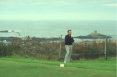}
   \caption{\footnotesize Inpainted by \cite{PGB}.}
  \label{golf:PGB}
    \end{subfigure}
\vskip0.2cm
    \begin{subfigure}[b]{0.235\textwidth}
        \includegraphics[width=\textwidth]{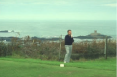}
   \caption{\footnotesize Inpainted by \cite{H}.}
  \label{golf:H}
    \end{subfigure}
   \hskip0.1cm
    \begin{subfigure}[b]{0.235\textwidth}
        \includegraphics[width=\textwidth]{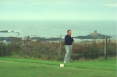}
   \caption{\footnotesize Inpainted by \cite{KT}.}
  \label{golf:PGB}
    \end{subfigure}
\vskip0.2cm
    \begin{subfigure}[b]{0.235\textwidth}
        \includegraphics[width=\textwidth]{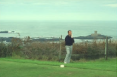}
   \caption{\footnotesize Inpainted by \cite{AFCS}  NL-means.}
  \label{golf:NLM}
    \end{subfigure}
   \hskip0.1cm
    \begin{subfigure}[b]{0.235\textwidth}
        \includegraphics[width=\textwidth]{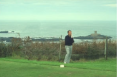}
   \caption{\footnotesize Inpainted by \cite{AFCS} NL-medians.}
  \label{golf:NLMd}
    \end{subfigure}
\vskip0.2cm
    \begin{subfigure}[b]{0.235\textwidth}
        \includegraphics[width=\textwidth]{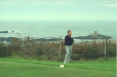}
   \caption{\footnotesize Inpainted by \cite{AFCS}  NL-Poisson.}
  \label{golf:NLP}
    \end{subfigure}
   \hskip0.1cm
    \begin{subfigure}[b]{0.235\textwidth}
        \includegraphics[width=\textwidth]{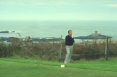}
   \caption{\footnotesize Inpainted by \cite{AFCS} NL-GM.}
  \label{golf:NLGM}
    \end{subfigure}
\vskip0.2cm
    \begin{subfigure}[b]{0.235\textwidth}
        \includegraphics[width=\textwidth]{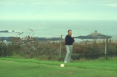}
   \caption{\footnotesize Inpainted by our method ($2L+1=3$, $l=1$).}
  \label{golf:our}
    \end{subfigure}
\end{center}
\caption{\footnotesize Inpainting by different well known methods in the literature. Images from (c) to (h) taken from \cite{AFCS}, with $2L+1>3$. In our method $2L+1=3$.}
\label{fig:golf} 
\end{figure}

\begin{figure}[!htb]
\begin{center}
    \begin{subfigure}[b]{0.235\textwidth}
        \includegraphics[width=\textwidth]{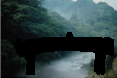}
   \caption{\footnotesize To be inpainted.}
  \label{bridge:To be inpainted}
    \end{subfigure}
   \hskip0.1cm
    \begin{subfigure}[b]{0.235\textwidth}
        \includegraphics[width=\textwidth]{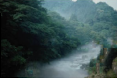}
   \caption{\footnotesize Inpainted by \cite{PGB}.}
  \label{bridge:PGB}
    \end{subfigure}
\vskip0.2cm
    \begin{subfigure}[b]{0.235\textwidth}
        \includegraphics[width=\textwidth]{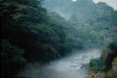}
   \caption{\footnotesize Inpainted by \cite{H}.}
  \label{bridge:H}
    \end{subfigure}
   \hskip0.1cm
    \begin{subfigure}[b]{0.235\textwidth}
        \includegraphics[width=\textwidth]{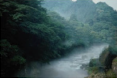}
   \caption{\footnotesize Inpainted by \cite{KT}.}
  \label{bridge:PGB}
    \end{subfigure}
\vskip0.2cm
    \begin{subfigure}[b]{0.235\textwidth}
        \includegraphics[width=\textwidth]{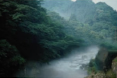}
   \caption{\footnotesize Inpainted by \cite{AFCS}  NL-means.}
  \label{bridge:NLM}
    \end{subfigure}
   \hskip0.1cm
    \begin{subfigure}[b]{0.235\textwidth}
        \includegraphics[width=\textwidth]{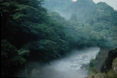}
   \caption{\footnotesize Inpainted by \cite{AFCS} NL-medians.}
  \label{bridge:NLMd}
    \end{subfigure}
\vskip0.2cm
    \begin{subfigure}[b]{0.235\textwidth}
        \includegraphics[width=\textwidth]{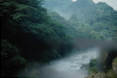}
   \caption{\footnotesize Inpainted by \cite{AFCS}  NL-Poisson.}
  \label{bridge:NLP}
    \end{subfigure}
   \hskip0.1cm
    \begin{subfigure}[b]{0.235\textwidth}
        \includegraphics[width=\textwidth]{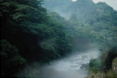}
   \caption{\footnotesize Inpainted by \cite{AFCS} NL-GM.}
  \label{bridge:NLGM}
    \end{subfigure}
\vskip0.2cm
    \begin{subfigure}[b]{0.235\textwidth}
        \includegraphics[width=\textwidth]{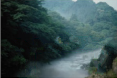}
   \caption{\footnotesize Inpainted by our method ($2L+1=3$, $l=1$).}
  \label{bridge:our}
    \end{subfigure}
\end{center}
\caption{\footnotesize Inpainting by different well known methods in the literature. Images from (c) to (h) taken from \cite{AFCS}, with $2L+1>3$. In our method $2L+1=3$.}
\label{fig:bridge} 
\end{figure}

\section{Conclusions}\label{Conclusions} 

In this paper a new functional, strictly connected with the definition of Sobolev spaces, but formalized in discrete setting, has been introduced to investigate the inpainting problem. The usual $MSE$-like metric (Mean Square Error) has been replaced with a more precise one, in the Sobolev spaces sense, for the determination of the correct values to inpaint. 

	The achieved results show that the new functional, taking into account of finite differences, improves the quality of the reconstructions. 

	The improvement is from one hand theoretically predictable (functions being equal, or similar, for higher order derivatives "resemble" more each other), on the other hand it has been numerically proved,  given a suitable training set. 

	In particular, results have shown the capability of the new formulation to carry, in the reconstructed area, precise structural information, a part of which are dependent on the size and shape of the neighborhood, that strongly conditions the final quality of the reconstructions. An application proof of this behavior stands in the capability of the new method to reproduce and connect isophotes, even in complex cases, and not  necessarily privileging the propagation along straight lines. 

	Moreover, the new formulation plays two complementary roles: from one hand it speeds up the convergence toward a minimum value (in terms of steps needed to reach it); on the other hand it reduces the number of local minima, lowering the probability to end the process in one of them. 

	In addition, the role played by the causality condition has been investigated and considered to be a starting point in the direction of the uncertainty reduction of the entire process.

	Following this reduction, a new priority index has been formalized, giving an applicative explanation of the roles played by each one of its terms: the inpainting energy $E_T$, the trustability $C$ and the finite differences $D_M$. All of them concur together to the uncertainty reduction. 

	The numerous numerical results, both on synthetic as on natural images, confirm the expectations, as in terms of the mathematical model, as in terms of the values calculated for the parameters in the applications. The visual quality of the results overpass what achievable with some state of the art techniques. 

	The actual drawback of the proposed method, as it has been implemented without any optimization, resides in the execution time: the possibility of the  improvement of the efficiency and the consequent reduction of the numerical complexity, e.g., moving to the Fourier Transform domain as well as using methods similar to \cite{BSAD} or tree structures as TSVQ (\cite{WEI}), \textit{k}d-trees (\cite{HER,WSI,KFCDLW}), VP-trees (\cite{KZN}), will be the subject of future studies and implementations.

	In conclusion the new proposed approach reduces the uncertainty in the inpainting process, boosting the amount of information that can be inferred from the known part of the image data.

\appendix[List of symbols]

\begin{itemize}
\item $a$ row shift value in $\supp \chi$;
\item $a^*$ normalization coefficient for $P^*$;
\item $\alpha$ normalization coefficient for $D$;
\item $b$ column shift value of $\supp \chi$;
\item $b^*$ normalization coefficient for $P^*$;
\item $\beta_k$ normalization coefficient for $E_{S_{T}}$;
\item $C$ trustability or confidence term;
\item $\chi$ neighborhood function;
\item $D$ state of the art data term;
\item $D_M$ proposed new data term;
\item $d$ pointwise distance between two pixels in $I$;
\item $d_{CNorm}$ normalized pointwise content-related distance between two pixels in $I$;
\item $d_{SNorm}$ normalized pointwise structure-related distance between two pixels in $I$;
\item $\Delta_{\theta}^{(k)}$ finite difference operator of order $k$ in direction $\theta$;
\item $\partial \Omega$ boundary of $\Omega$;
\item $E_C$ single point, content-related, inpainting energy;
\item $E_{C_{T}}$ total content-related inpainting energy;
\item $E_I$ inpainting energy;
\item $E_{M_{C}}$ total patch content-related probability of a match in $T_S$ given $\supp \chi$;
\item $E_{M_{S}}$ total patch structure-related probability of a match in $T_S$ given $\supp \chi$;
\item $E_{P_{C}}$ pointwise content-related probability of a match in $T_S$ given $\supp \chi$;
\item $E_{P_{S}}$ pointwise structure-related probability of a match in $T_S$ given $\supp \chi$;
\item $E_S$ single point, structure-related, inpainting energy;
\item $E_{S_{T}}$ total structure-related inpainting energy;
\item $E_T$ total inpainting energy;
\item $I$ image under investigation;
\item $i$, $h$, $r$, $p$ row coordinate of a pixel in $I$;
\item $j$, $m$, $s$, $q$ column coordinate of a pixel in $I$;
\item $k$ generic order of the finite differences;
\item $l$ maximum order of the finite differences;
\item $\ell$ discrete Lebesgue space;
\item $M$ number of rows in $I$;
\item $N$ number of columns in $I$;
\item $n$ unitary vector orthogonal to $\partial \Omega$;
\item $\nabla^{\perp}$ isophote vector;
\item $\Omega$ to-be-inpainted area;
\item $P$ state of the art scanning priority index;
\item $P^*$ proposed scanning priority index;
\item $R_k$ direction-related normalization coefficient for $E_{S_{T}}$;
\item $\mathcal{S}$ specific configuration in $S_C$;
\item $S_C$ set of equivalence classes;
\item $\supp \chi$ support of the neighborhood function;
\item $t$ time;
\item $T_S$ training set;
\item $\theta$ specific orientation of the finite difference;
\item $\Theta_k$ set of possible and valid orientations $\theta$ given the order $k$;
\item $W$ bit depth;
\item $U$ uncertainty.
\end{itemize}

  \section*{Acknowledgments}

M. Seracini thanks prof. L. Demanet of MIT for introducing him the inpainting problem and prof. A. Fournier of University of Colorado for some interesting discussions about the topic. Moreover, M. Seracini thanks Unione Matematica Italiana (UMI) for its grant.

\end{document}